\documentclass[twoside,11pt]{article}

\usepackage{jmlr2e}

\usepackage{algorithmicx}
\usepackage[Algorithm,ruled]{algorithm}
\usepackage[noend]{algpseudocode}

\usepackage{color}
\usepackage{enumerate}
\usepackage{mathtools}
\usepackage{amsmath,amssymb}
\usepackage{footnote}
\makesavenoteenv{tabular}

\usepackage{environ}
\usepackage{tikz}
\usetikzlibrary{arrows.meta}
\usetikzlibrary{quotes}
\usetikzlibrary{positioning,automata,decorations.pathreplacing, decorations.markings, shapes.misc, shapes.geometric}

\usepackage{hyperref}
\hypersetup{
  colorlinks   = true,
  urlcolor     = blue,
  linkcolor    = blue,
  citecolor    = blue
}

\definecolor{new_red}{rgb}{.8,0,0}
\definecolor{new_blue}{rgb}{0,0,.8}
\definecolor{new_green}{rgb}{0,.5,0}

\newcommand{\Green}[1]{\textcolor{new_green}{#1}}
\newcommand{\Red}[1]{\textcolor{red}{#1}}
\newcommand{\Blue}[1]{\textcolor{new_blue}{#1}}
\newcommand{\Purple}[1]{\textcolor{purple}{#1}} 
\newcommand{\Violet}[1]{\textcolor{violet}{#1}}

\newenvironment{newproof}{ \par\noindent{\bfseries\upshape Proof.\ }}{\hfill $\blacksquare$}

\newcommand{\ignore}[1]{} 

\usepackage{tikz}
\usetikzlibrary{arrows,positioning, calc}

\usepackage{graphicx}
\usepackage{caption}
\usepackage{subcaption}
\usepackage{enumitem}

\usepackage{bbm}

\input{Definitions.tex}

\ShortHeadings{Online Dynamic Programming}{Rahmanian, Warmuth and Vishwanathan}
\firstpageno{1}

\begin{document}

\title{Online Dynamic Programming}

\author{%
\name Holakou Rahmanian \email holakou@amazon.com\\
\addr Amazon\\
\name Manfred K. Warmuth \email manfredwarmuth57@gmail.com\\
\addr Google Research\\
\name S.V.N. Vishwanathan \email vishy@amazon.com\\
\addr Amazon
}


\maketitle

\begin{abstract}
We propose a general method for combinatorial online learning problems whose offline optimization problem can be solved efficiently 
via a dynamic programming algorithm defined by an arbitrary min-sum recurrence.
Examples include online learning of
Binary Search Trees, Matrix-Chain Multiplications, $k$-sets, Knapsacks, Rod Cuttings, and Weighted Interval Schedulings.
For each of these problems we use
the underlying graph of subproblems (called a \emph{multi-DAG}) 
for defining a representation of the 
solutions of the dynamic programming problem by encoding them 
as a generalized version of paths (called \emph{multipaths}).
These multipaths encode each solution as a series of successive decisions 
or components over which the loss is linear. 
We then show that the dynamic programming algorithm for each problem
leads to online algorithms for learning multipaths in the underlying multi-DAG. 
The algorithms maintain a distribution over the multipaths in a concise form as their hypothesis.
More specifically we generalize the existing
Expanded Hedge \citep{takimoto2003path} and Component
Hedge  \citep{koolen2010hedging} algorithms for the online shortest path problem to learning multipaths.
Additionally, we introduce a new and faster prediction technique for 
Component Hedge which in our case directly samples from a distribution over multipaths, 
bypassing the need to decompose the distribution over
multipaths into a mixture with small support.
\end{abstract}

\section{Introduction}
\label{sec:introduction}

We consider the problem of repeatedly solving the same
dynamic programming problem in successive trials. The set
of subproblems must remain unchanged but the losses/costs of the
solutions change in each trial.
We can handle any dynamic programming problem specified by 
arbitrary min-sum recurrence relations.
The original well-studied
problem of this type is to learn paths in a given fixed directed acyclic graph (DAG)
with designated source and sink nodes.
In this case, the minimum loss paths are related by rudimentary min sum recurrence and
the loss of each path from the source to the sink is \emph{additive}, that is,
it is the sum of the losses of the edges along that path.
For the explanation purposes, we start with this path learning problem
and will introduce the online learning setup with this example problem.
However this will be generalized later to handle dynamic programming
problems defined by arbitrary min-sum recurrence relations.

The path learning problem
is specified by a DAG $G=(V,E)$.
For every node $v\in V$ in the given DAG, 
we let OPT$(v)$ denote the loss of the best path from $v$ to the sink.
We have the following min-sum recurrence relation:
$$
\text{OPT}(v) = \Red{\min_{u: (v,u) \in E}} \{ \text{OPT}(u) \Red{+} \ell_{(v,u)} \},
$$
where $\ell_e$ is the loss of the edge $e \in E$.

Online learning of paths in $G$ 
proceeds in a series of trial.
In each trial, the \emph{learner} predicts with a path in $G$.
Then, the \emph{adversary} reveals the losses of all the edges in $E$.
Finally, the learner incurs the loss of its predicted path.
The goal is to minimize \emph{regret} which is the total loss of the learner minus the total loss of the single best path in hindsight.

A natural appraoch is to use the well-known ``expert algorithms'' like 
\emph{Randomized Weighted Majority} \citep{littlestone1994weighted} or \emph{Hedge}
\citep{freund1997decision} with the paths serving as the experts.
These algorithms maintain one weight per path (of which there are exponentially many).
However, exploiting the additivity of the loss,
\cite{takimoto2003path} gave an efficient implementation of
the Hedge algorithm for the path learning problem (called \emph{Expanded Hedge (EH)}).
EH assigns weights to the edges and implicitly maintains a distribution over paths 
where the probability of each path is proportional 
to the product of the weights of the edges along that path.
Another efficient algorithm for learning paths is 
the \emph{Component Hedge (CH)} algorithm of \cite{koolen2010hedging}
which is a generic algorithm for combinatorial online learning
with additive losses over the components (here the edges).
Instead of a distribution, CH maintains a mean vector over the paths and assigns \emph{flows} to the edges.
This mean vector lies in the \emph{unit-flow polytope} which is the convex hull of all paths in the graph.
Comparing to EH, 
CH guarantees better regret bounds as it does not have
maximum loss of the paths in its bounds.
We will expand on this in Section~\ref{sec:ch-vs-eh-paths}.

In this paper we generalize the online shortest path problem to
learning any min-sum dynamic programming problem.
The set of subproblems is fixed between trails.
In each trial, the learner predicts with a solution.
It then receives the losses of all solutions (by receiving the losses of all components).
Note that there is no assumed distribution from which the losses are drawn. 
The goal is to minimize \emph{regret} 
(the total loss of the learner minus the total loss of the single best solution in hindsight)
for any (possibly adversarial) sequence of losses between trials.
With proper tuning of the algorithms, 
the regret is typically logarithmic in the number of solutions.

For sake of concreteness, consider the problem of learning the best
\underline{B}inary \underline{S}earch \underline{T}ree (BST) for a given fixed set of $n$ keys \citep{thomas2001introduction}.
In each trial, the learner plays with a BST.
Then the adversary reveals a set of probabilities for the $n$ keys 
and the learner incurs a linear loss of \textit{average search cost}, 
which is simply the dot product between the vector of probabilities and the vector of depth values of the keys 
in the tree.
The regret of the learner is the difference between its total loss and the sum over trials of the average search cost for the single best BST chosen in hindsight.

The number of solutions is typically exponential in $n$ 
where $n$ is the number of components in the structure of the solutions.
In a BST, the components are the depth values of the $n$ keys 
in the tree, and the number of possible BSTs is the 
$n$th Catalan number $C_n = \frac{1}{n+1} {2n \choose n}$ \citep{thomas2001introduction}.
Thus as for the path problem, naive implementations of Randomized Weighted Majority or Hedge 
(i.e\ maintaining one weight per BST) is inefficient, 
and in this paper, we extent the EH implementation to
handle all problems with min-sum recurrence relations.

Also, the CH algorithm (and its current extensions \citep{suehiro2012online, rajkumar2014online, gupta2016solving}) 
cannot be directly applied to problems like BST.
The reason is that CH maintains a mean vector of the BSTs 
which lives in the convex hull of all BSTs with the representation above
and this polytope does not have a characterization with polynomially many facets%
\footnote{There is an alternate polytope for BSTs with a polynomial number of facets (called the \emph{associahedron} \citep{loday2005multiple})
but the average search cost is not linear in terms of the components used for this polytope.
CH and its extensions, however, rely heavily on the additivity of the loss over the components.
Thus they cannot be applied to the associahedron.
}.
%

In this paper we manage to construct an alternate
representation from the offline dynamic programming
algorithm for say the BST problem.
This gives us a polytope with a polynomial number of facets while the loss is linear in the natural components of the BST problem.
This well-behaved polytope will allow us to implement CH efficiently.
We also use this representation to implement EH efficiently.

\begin{table}
\centering
\begin{tabular}{ c  c  c}
\textbf{Graph} & $\Longrightarrow$ & \textbf{Multigraph} \\
with the set of vertices $V$ & ~ & with the set of vertices $V$ \\
\hline
\textbf{Edge} $(v,u)$ & ~ & \textbf{Multiedge} $(v,U)$ \\
$u,v \in V$ & $\Longrightarrow$ &  $v \in V, \; U \subset V$\\
\scalebox{.5}{
\begin{tikzpicture}%
  [>=stealth,
   shorten >=1pt,
   node distance=2cm,
   on grid,
   auto,
   every state/.style={draw=black!60, fill=black!5, very thick}
  ]
\node[state] (vone)                  {$v$};
\node[state] (u) [right=of vone] {$u$};

\path[->]
	(vone) edge node { } (u)
   ;
   
\node { }  (0,-2.5);
\end{tikzpicture}
}
&~ &
\scalebox{.5}{
\begin{tikzpicture}%
  [>=stealth,
   shorten >=1pt,
   node distance=2cm,
   on grid,
   auto,
   every state/.style={draw=black!60, fill=black!5, very thick}
  ]
\node[state] (vtwo) {$v$};
\node[state] (utwo) [right=of vtwo] {$\vdots$};
\node[state] (uone) [above=of utwo] {$u_1$};
\node[state] (uthree) [below=of utwo] {$u_k$};

\path[->]
	(vtwo) edge node { } (uone)
	(vtwo) edge node { } (utwo)
	(vtwo) edge node { } (uthree)
   ;
\end{tikzpicture}
}
\\
\hline
\textbf{Path} &  $\Longrightarrow$ & \textbf{Multipath} \\
\hline
\textbf{DAG} &  $\Longrightarrow$ & \textbf{Multi-DAG} \\
\end{tabular}
\caption{From graphs to multi-graphs}
\label{table:multigraph}
\end{table}

\paragraph{Paper Outline.}
We start with online learning of paths in a directed graph in Section~\ref{sec:odp-background}
and give an overview on existing algorithms: Expanded Hedge and Component Hedge.
In Section~\ref{sec:multipaths}, 
We generalize the definition of edge as an ordered pair $(v,u)$ of vertices to  
\emph{multiedge} which is an ordered pair $(v,U)$ where the first element $v$ is a vertex
and the second element $U$ is a \emph{subset of the vertices}.
Following from this generalization, we extend the definitions of paths, graphs and directed acyclic graphs (DAG) accordingly 
(see Table~\ref{table:multigraph}; Section~\ref{sec:multipaths} contains the formal definitions).
These extensions allow us to generalize the existing EH (Section~\ref{sec:eh-multipaths}) and CH (Section~\ref{sec:ch-multipaths}) 
algorithms from online shortest path problem to learning multipaths.
For EH, we introduce the \emph{stochastic product form} as a concise way of maintaining a distributions over all multipaths. 
For CH, on the other hand, we maintain the mean vector of a distribution in the \emph{unit-flow polytope} which has polynomial number of facets.
Moreover, we also introduce a new and faster prediction technique for CH for multipaths which directly samples from an appropriate distribution, 
bypassing the need to create convex combinations.
More specifically, 
we use the maintained mean vector in the unit-flow polytope in the CH algorithm
and construct an appropriate distribution in the stochastic product form.
In Section~\ref{sec:dp-games}, we define a general class of combinatorial online learning problems 
which can be recognized by dynamic programming algorithms. 
Then we prove that minimizing a specific dynamic programming problem from this class over trials reduces to online learning of multipaths. 
In Section~\ref{sec:instantiation}, we apply our methods to several dynamic programming problems.
Finally, Section~\ref{sec:odp-conclusions} concludes with comparison to other algorithms and future work.

\section{Background}
\label{sec:odp-background}

One of the core combinatorial online learning problems is learning a minimum loss path in a directed acyclic graph (DAG).
The online shortest path problem has been explored both 
in the full information setting \citep{takimoto2003path, koolen2010hedging, cortes2015line}
and various bandit settings \citep{gyorgy2007line, audibert2013regret, awerbuch2008online, dani2008price, cortes2018nab}. 
In the full information setting, the problem is as follows.
A DAG $\Gcal = (V,E)$ is given along with a designated source node $s \in V$ and sink node $t \in V$. 
In each trial, the algorithm predicts with a path from $s$ to $t$. 
Then for each edge $e \in E$, the adversary reveals a loss $\ell_e \in [0,1]$. 
The loss of the algorithm is given by the sum of the losses of the edges (components) along the predicted path. 
The goal is to minimize the regret which is the difference between the total loss of the algorithm and that of the single best path chosen in hindsight.
%
In the remainder of this section, we 
provide overviews of EH and CH as the the two main algorithms for online path learning in full information setting.

\subsection{Expanded Hedge on Paths}
\label{sec:eh-paths}
\cite{takimoto2003path} introduced Expanded Hedge (EH) algorithm for online path learning problem.
EH is an efficient implementation of 
the Hedge algorithm which exploits the additivity of the loss over the edges of a path.
%
%
Viewing each path as an ``expert'', 
the weight $w_\pi$ of a path $\pi$ is proportional to $\prod_{e\in\pi} \exp(-\eta L_e)$, 
where $L_e$ is the cumulative loss of edge $e$.
The algorithm maintains one weight $w_e$ per edge $e\in E$.
These weights are in \emph{stochastic form}, that is, the total weight of all edges leaving any non-sink node sums up to $1$.
The weight of each path is in \emph{product form} $w_\pi=\prod_{e\in\pi} w_e$ and sampling a path is easy.
At the end of the current trial, each edge $e$ receives additional loss $\ell_e$, and path weights are updated.
The multiplicative updates with exponentiated loss for the paths decomposes over the edges due to additivity of the loss over the edges.
Thus the updated path weights will be
$$
w^\n_\pi
=\frac{1}{Z} w_\pi \exp(-\eta \sum_{e \in \pi} \ell_e)
=\frac{1}{Z} \prod_{e\in\pi} w_e\exp(-\eta \ell_e),
$$
where $Z$ is the normalization. 
Now a certain efficient procedure called
\emph{weight pushing} \citep{mohri2009weighted} 
is applied. 
It finds new edge weights $w^\n_e$ which are again in stochastic product form,
i.e. the out-going weights at each node sum up to one 
and the updated weights are $w^\n_\pi=\prod_{e\in\pi} w^\n_e$, facilitating sampling.
EH provides the regret guarantees below.

\begin{theorem}[\cite{takimoto2003path}]
\label{thm:EH-paths}
Given a DAG $\Gcal = (V,E)$ with designated source node $s \in V$ and sink node $t \in V$, 
assume $\Ncal$ is the number of paths in $\Gcal$ from $s$ to $t$, 
$L^*$  is the total loss of best path,
and $D$ is an upper-bound on the number of edges of the paths in $\Gcal$ from $s$ to $t$.
Then with proper tuning of the learning rate $\eta$ over the trials, EH guarantees the following regret bound:
\begin{equation}
\mathcal{R}_\text{EH} \leq \sqrt{2 \, L^* \, D \, \log \Ncal} + D \, \log \Ncal.
\label{eq:EH_regret}
\end{equation}

\end{theorem}

\subsection{Component Hedge on Paths}
\label{sec:ch-paths}
The generic Component Hedge algorithm of \cite{koolen2010hedging} can be applied to the online shortest path problem. 
The components are the edges $E$ in the DAG.
Each path is encoded as a bit vector $\pivec$ of $|E|$ components where the $1$-bits indicate the presence of the edges in the path $\pivec$.  
The convex hull of all paths is called the \textit{unit-flow polytope} and CH maintains a mixture vector $\fvec=[f_e]_{e\in E}$ in this polytope.
The constraints of the polytope enforce an outflow of $1$ from the source node $s$,
and flow conservation at every other node but the sink node $t$.
In each trial, each edge $e$ receives a loss of $\ell_e$ and the weight of that edge $f_e$ is updated multiplicatively by the factor $\exp(-\eta\ell_e)$.
Then the weight vector is projected back to the unit-flow polytope via a relative entropy projection. 
This projection is achieved by iteratively projecting onto the flow constraint of a particular vertex 
and then repeatedly cycling through the vertices \citep{bregman1967relaxation}.
Finally, to sample with the same expectation as the mixture vector in the polytope, 
this vector is decomposed into paths using a greedy approach 
which removes one path at a time 
and zeros out at least one edge in the remaining mixture vector in each iteration. 
CH provides the regret guarantees below.

\begin{theorem}[\cite{koolen2010hedging}]
\label{thm:CH-paths}
Given a DAG $\Gcal= (V,E)$ with designated source node $s \in V$ and sink node $t \in V$, 
let $D$ be an upper-bound on the number of edges of the paths in $\Gcal$ from $s$ to $t$.
Also denote the total loss of the best path by $L^*$. 
Then with proper tuning of the learning rate $\eta$ over the trials, CH  guarantees the following regret bound:
\begin{equation}
\mathcal{R}_\text{CH} \leq \sqrt{4 \, L^* \, D \, \log |V|} + 2 \, D \log |V|.
\label{eq:CH_regret}
\end{equation}
\end{theorem}

\paragraph{Remark.}
In a moment, we will compare the regret bounds of EH (\ref{eq:EH_regret}) and CH (\ref{eq:CH_regret}).
We will observe that 
compared to EH, CH guarantees better regret bounds as it does not have additional loss range factors in its bounds.
In fact, the regret bounds of CH are typically optimal. 
\cite{koolen2010hedging} prove lower bounds which matches the guarantees of CH for various 
problems such as $k$-sets and permutations.
The lower bounds are shown by embedding the combinatorial online learning into the original expert problem.
To form the experts, a set of solutions is chosen which partitions all of the components.
Moreover, all solutions in the set must have the same number of present components (i.e.~same number of ones in the bit-vector representation).
Given this proof technique, a lower bound on the regret for arbitrary graphs is difficult to obtain
since choosing a set of paths with the aforementioned characteristics is non-trivial.
Perhaps the regret of CH is tight within constant factors for all graphs, but this question is still open.

\subsection{Component Hedge vs Expanded Hedge}
\label{sec:ch-vs-eh-paths}
To have a concrete comparison between CH and EH on paths, consider the following path learning setting. 
Let $\Gcal = (V, E)$ be a complete DAG with $V = \{ v_1, \ldots, v_n \}$ where for all $1\leq i<j \leq n$, $v_i$ is connected to $v_j$. 
Also let $s=v_1$ and $t=v_n$ be the designated source and sink nodes, respectively.
Note that the number of edges in any path in $\Gcal$ from $s$ to $t$ is at most $D=n-1$.
Also the total number of paths in $\Gcal$ is $\Ncal = 2^{n-2}$.
The corollary below shows the superiority of the performance of CH over EH in terms of regret bound which is 
a direct result of Theorems~\ref{thm:EH-paths} and~\ref{thm:CH-paths}.
EH offers worse regret guarantee as its bound has an additional loss range factor. 

\begin{corollary}
\label{crlry:CH-vs-EH}
Given a complete DAG $\Gcal$ with $n$ nodes, let $L^*$ be the total loss of the best path.
Then with proper tuning of the learning rate $\eta$ over the trials for both EH and CH,
we obtain the following regret guarantees:
$$
\mathcal{R}_\text{EH} = \Ocal(n \sqrt{L^*}),
\qquad
\mathcal{R}_\text{CH} = \Ocal(n^{\frac{1}{2}} \, (\log n)^{\frac{1}{2}} \sqrt{L^*}).
$$
\end{corollary}

\paragraph{Remark.} 
For EH, projections are simply a renormalization of the path weights which is done efficiently via weight pushing \citep{mohri2009weighted, takimoto2003path}.
On the other hand, for CH, iterative Bregman projections \citep{bregman1967relaxation} are needed
for projecting back into the unit-flow \citep{koolen2010hedging}.
These methods are known to converge to the exact projection \citep{bregman1967relaxation, bauschke1997legendre}; 
however, there will always be a gap to the full convergence.
These remaining gaps to the exact projections have to be accounted for as additional loss in the regret bounds (e.g.~see \cite{rahmanian2018xfhedge}).
Additionally, the relatively expensive projection operation in CH makes it less computationally efficient compared to EH.

\section{Learning Multipaths} 
\label{sec:multipaths}

We begin with defining directed multigraphs, multiedges%
\footnote{
Our definitions of multigraphs and multiedges are closely related to 
\emph{hyper-graphs} and \emph{hyper-arcs} in the literature (see e.g.\ \cite{martin1990polyhedral}).
} and multi-DAGs.

\begin{definition}[Directed Multigraph]
\label{def:multigraph}
A \textbf{directed multigraph} is an ordered pair $\Hcal = (V, M)$ comprising of a set $V$ of vertices or nodes together with a set $M$ of \textbf{multiedges}. 
Each multiedge $m \in M$ is an ordered pair $m = (v, U)$ where $v \in V$ and $U \subseteq V$. 
Furthermore, we denote the set of ``outgoing'' and ``incoming'' multiedges for vertex $v$ by $M^\text{(out)}_v$ and $M^\text{(in)}_v$, respectively,
which are defined as 
\begin{align*}
&M^\text{(out)}_v := \{ m \in M \mid m=(v,U) \text{ for some } U \subseteq V \},\\
&M^\text{(in)}_v := \{ m \in M \mid m=(u,U) \text{ for some } u \in V \text{ and } U \subseteq V \text{ s.t. } v \in U \}.
\end{align*}
\end{definition}

\begin{definition}[Base Directed Graph]
\label{def:basegraph}
The \textbf{base directed graph} of a given directed multigraph $\Hcal = (V, M)$ is a directed graph $\Bcal(\Hcal) = (V, E)$ where
$$
E = \{ (v,u) \mid \exists (v,U) \in M \text{ s.t. } u \in U \}.
$$
\end{definition}

\begin{definition}[Multi-DAG]
\label{def:multidag}
A directed multigraph $\Hcal = (V, M)$ is a \textbf{multi-DAG} 
if it has a single ``source'' node $s \in V$ with no incoming multiedges and
its base directed graph $\Bcal(\Hcal)$ is acyclic.
Additionally, we refer to the set of nodes in $V$ with no outgoing multiedges 
as the set of ``sink'' nodes which is denoted by $\Tcal \subset V$.
\end{definition}

Intuitively speaking, a multi-DAG is simply a directed multigraph with no ``cycles''.
``Acyclicity'' in directed multigraphs%
\footnote{
For our application of dynamic programming,
this acyclicity is very natural; otherwise the dynamic programming algorithm is 
not valid and a subproblem may be visited infinite number of times.
} %
can be extended from the definition of acyclicity in the underlying directed graph.

Each multipath in a multi-DAG $\Hcal = (V, M)$ can be generated by starting with a single multiedge at the source, 
and then choosing inflow many (i.e.~as many as the number of incoming edges of the multipath in $\Bcal(\Hcal)$) 
successor multiedges at the internal nodes until we reach the sink nodes in $\Tcal$.
An example of a multipath is given in Figure~\ref{fig:multidag}.
Recall that paths were described as bit vectors $\pivec$ of size $|E|$ where the $1$-bits were the edges in the path. 
In multipaths, however, each multiedge $m \in M$ is associated with a non-negative count $\pi_m$
that can be greater than $1$.

\begin{definition}[Multipath]
\label{def:multipath}
Given a multi-DAG $\Hcal = (V,M)$, let\footnote{$\NN$ is the set of non-negative integers.} 
 $\pivec \in \NN^{|M|}$ in which $\pi_m$ is associated with the multiedge $m \in M$. 
For every vertex $v \in V$, define the inflow $\pi_\text{in}(v) := \sum_{m \in M^\text{(in)}_v} \pi_m$ 
and the outflow $\pi_\text{out}(v) := \sum_{m \in M^\text{(out)}_v} \pi_m$.
We call $\pivec$ a \textbf{multipath} if it has the properties below:
\begin{enumerate}
\item The outflow $\pi_\text{out}(s)$ of the source $s$ is $1$.
\item For each vertex $v \in V\!-\!\Tcal\!-\!\{s\}$, 
the outflow is equal to the inflow, i.e. $\pi_\text{out}(v) = \pi_\text{in}(v)$.
\end{enumerate}
\end{definition}


\begin{figure}
\centering
\begin{minipage}[R]{0.5\textwidth}
\scalebox{.9}{
\begin{tikzpicture}%
  [>=stealth,
   shorten >=1pt,
   node distance=1.5cm,
   on grid,
   auto,
   every state/.style={draw=black, fill=gray!15, minimum size=10pt}
  ]
  
\tikzset{rectangle/.style={draw=gray,dashed,thick,inner sep=5pt}}
\tikzset{bedge/.style = {->,> = stealth,  thick, blue}}
\tikzset{redge/.style = {->,> = stealth,  thick, red}}
\tikzset{gedge/.style = {->,> = stealth,  thick, new_green}}

\node[rectangle, anchor=south west, minimum width=1cm, minimum height=2.5cm] (A) at (5.5,-1.4) {};
\node[text width=1cm] at (6.3, 1.8) {\large $\Tcal$};
\node[text width=1cm] at (.4,1) {\large $s$};

\node[state] (0) { } ;

\node[state] (12) [above right=.65cm and 2cm of 0] { } ;
\node[state] (13) [below=of 12] { } ;
\node[state] (11) [above=of 12] { } ;
\node[state] (14) [below=of 13] { } ;

\node[state] (21) [right=2cm of 11] { } ;
\node[state] (22) [right=2cm of 12] { } ;
\node[state] (23) [right=2cm of 13] { } ;
\node[state] (24) [right=2cm of 14] { } ;

\node[state] (32) [right=2cm of 22] { } ;
\node[state] (33) [right=2cm of 23] { } ;

\draw[bedge] (0) to (13);
\draw[bedge] (0) to (14);
\draw[redge] (0) to (11);
\draw[redge] (0) to (12);

\draw[bedge] (11) to (21);
\draw[bedge] (11) to (22);
\draw[redge] (11) to (23);
\draw[redge] (11) to (24);

\draw[bedge] (12) to (21);
\draw[redge] (12) to (22);
\draw[bedge] (12) to (23);
\draw[redge] (12) to (24);

\draw[redge] (13) to (21);
\draw[bedge] (13) to (22);
\draw[bedge] (13) to (23);
\draw[redge] (13) to (24);

\draw[redge] (14) to (21);
\draw[redge] (14) to (22);
\draw[bedge] (14) to (23);
\draw[bedge] (14) to (24);


\draw[gedge] (21) to (32);
\draw[gedge] (21) to (33);

\draw[gedge] (22) to (32);
\draw[gedge] (22) to (33);

\draw[gedge] (23) to (32);
\draw[gedge] (23) to (33);

\draw[gedge] (24) to (32);
\draw[gedge] (24) to (33);

\path[->]
   ;
   

\end{tikzpicture}
}
\end{minipage}
\hfill
\begin{minipage}[R]{0.4\textwidth}
\scalebox{.9}{
\begin{tikzpicture}%
  [>=stealth,
   shorten >=1pt,
   node distance=1.5cm,
   on grid,
   auto,
   every state/.style={draw=black, fill=gray!15, minimum size=10pt}
  ]
  
\tikzset{rectangle/.style={draw=gray,dashed,thick,inner sep=5pt}}
\tikzset{bedge/.style = {->,> = stealth, very thick, black}}
\tikzset{gedge/.style = {->,> = stealth,  thick, gray!20}}

\node[rectangle, anchor=south west, minimum width=1cm, minimum height=2.5cm] (A) at (5.5,-1.4) {};
\node[text width=1cm] at (6.3, 1.8) {\large $\Tcal$};
\node[text width=1cm] at (.4,1) {\large $s$};

\node[state] (0) { } ;

\node[state] (12) [above right=.65cm and 2cm of 0] { } ;
\node[state] (13) [below=of 12] { } ;
\node[state] (11) [above=of 12] { } ;
\node[state] (14) [below=of 13] { } ;

\node[state] (21) [right=2cm of 11] { } ;
\node[state] (22) [right=2cm of 12] { } ;
\node[state] (23) [right=2cm of 13] { } ;
\node[state] (24) [right=2cm of 14] { } ;

\node[state] (32) [right=2cm of 22] { } ;
\node[state] (33) [right=2cm of 23] { } ;

\draw[bedge] (0) to  (13) ;
\draw[bedge] (0) to node[black, text width = 1cm, near start, below] {$1$} (14);
\draw[gedge] (0) to (11);
\draw[gedge] (0) to (12);

\draw[gedge] (11) to (21);
\draw[gedge] (11) to (22);
\draw[gedge] (11) to (23);
\draw[gedge] (11) to (24);

\draw[gedge] (12) to (21);
\draw[gedge] (12) to (22);
\draw[gedge] (12) to (23);
\draw[gedge] (12) to (24);

\draw[gedge] (13) to (21);
\draw[bedge] (13) to node[black, text width = 1cm, above, near start] {$1$} (22);
\draw[bedge] (13) to (23);
\draw[gedge] (13) to (24);

\draw[bedge] (14) to (21);
\draw[bedge] (14) to node[black, text width = 1cm, right, near start] {$1$}(22);
\draw[gedge] (14) to (23);
\draw[gedge] (14) to (24);


\draw[bedge] (21) to node[black, text width = 1cm, right, near start] {$1$} (32);
\draw[bedge] (21) to (33);

\draw[bedge] (22) to (32);
\draw[bedge] (22) to node[black, text width = 1cm, below, near start] {$2$} (33);

\draw[bedge] (23) to (32);
\draw[bedge] (23) to node[black, text width = 1cm, below] {$1$} (33);

\draw[gedge] (24) to (32);
\draw[gedge] (24) to (33);

\path[->]
   ;
   

\end{tikzpicture}
}
\end{minipage}
\caption[Examples of multi-DAGs and multipaths.]
{ 
On the left we give an example of a multi-DAG. 
The source $s$ and the nodes in the first layer each
have two multiedges depicted in \textcolor{red}{red} and 
\textcolor{blue}{blue}.
The nodes in the next layer each have one multiedge
depicted in \textcolor{new_green}{green}.
An example of multipath in the multi-DAG is given on the
right. The multipath is represented as
an $|M|$-dimensional count vector $\pivec$.
The grayed multiedges are the ones with count $\pi_e=0$.
All non-zero counts $\pi_m$ are shown next to their
associated multiedges $m$.
}
\label{fig:multidag}
\end{figure}

\paragraph{Multipath Learning Problem.}
Having established all definitions for multipaths, 
we shall now define the problem of \textit{online learning of multipaths} on a given multi-DAG $\Hcal = (V, M)$ as follows.
In each trial, the algorithm randomly predicts with a multipath $\pivec$. 
Then for each multiedge $m \in M$, the adversary reveals a loss $\ell_m \in [0,1]$ incurred during that trial. 
The linear loss of the algorithm during this trial is given by $\pivec\cdot \ellvec$. 
Observe that the online shortest path problem is a special case when $|\Tcal|=1$ and $|U|=1$ for all multiedges $(v,U) \in M$.

In the remainder of this section, we generalize the algorithms in Section~\ref{sec:odp-background} to the online learning problem of multipaths.
Moreover, we also introduce a faster prediction technique for CH.

\subsection{Expanded Hedge on Multipaths}
\label{sec:eh-multipaths}

We implement EH efficiently for learning multipaths by considering each multipath as an expert.
Recall that each multipath can be generated by starting with a single multiedge at the source 
and choosing inflow many successor multiedges at the internal nodes.
Multipaths are composed of multiedges as components and with each multiedge $m \in M$, we associate a weight $w_{m}$.
We maintain a distribution $W$ over multipaths defined in terms of the weights $\wvec \in \RR^{|M|}_{\geq 0}$ on the multiedges. 
The distribution $W$ will be in \emph{stochastic product form} which is defied as below.

\begin{definition}[Stochastic Product Form] 
\label{def:stochastic}
The distribution $W$ over the multipaths is in \textbf{stochastic product form} in terms of the weights $\wvec$ if it has the following properties:
\begin{enumerate}
\item The weights are in \textbf{product form}, i.e.~ $W(\pivec) = \prod_{m \in M} (w_{m})^{\pi_{m}}$.
\item The weights are \textbf{stochastic}, i.e.~ $\sum_{m \in M^\text{(out)}_v} w_{m}=1$ for all $v \in V \!-\!\Tcal$.
\item The total path weight is one%
\footnote{
The third property is implied by the first two properties. 
Nevertheless, it is mentioned for the sake of clarity.
}
, i.e. $\sum_{\pivec} W(\pivec) = 1$.
\end{enumerate}
\end{definition}

Using these properties, sampling a multipath from $W$ can be easily done as follows. 
We start with sampling a single multiedge at the source 
and continue sampling inflow many successor multiedges at the internal nodes
until the multipath reaches the sink nodes in $\Tcal$.
Observe that $\pi_{m}$ indicates the number of times the multiedge $m$ is sampled through this process. 
EH updates the weights of the multipaths as follows:
\begin{align*}
W^\n(\pivec) &=\frac{1}{Z} W(\pivec) \, \exp( - \eta \, \pivec \cdot \ellvec) \\
&=\frac{1}{Z}\left( \prod_{m \in M} (w_{m})^{\pi_{m}} \right)\, 
\exp \left[ - \eta \,  \sum_{m \in M} \pi_{m} \, \ell_m   \right] \\
&= \frac{1}{Z}\prod_{m \in M} \Big( 
\underbrace{w_{m}  \, \exp \Big[ - \eta \,   \ell_m \Big]}_{:=\what_{m}} \Big)^{\pi_{m}}.
\end{align*}

Thus the weights $w_m$ of each multiedge $m \in M$ are updated multiplicatively to $\what_m$ 
by multiplying the $w_m$ with the exponentiated loss factors $\exp \left[ - \eta \, \ell_m \right]$
and then renormalizing with $Z$.

\paragraph{Generalized Weight Pushing.}

We generalize the \textit{weight pushing algorithm} of \cite{mohri2009weighted} to multipaths 
to reestablish the three canonical properties of Definition~\ref{def:stochastic}.
Observe that for every multiedge $m \in M$, $\what_m = w_m \exp( - \eta \ell_m)$. 
The new weights are $W^\n(\pivec) = \frac{1}{Z} \What(\pivec)$ 
where $\What(\pivec) := \prod_{m \in M} (\what_{m})^{\pi_{m}}$.
The generalized weight pushing algorithm takes a set of arbitrary weights on 
the multiedges $\what_m$ and changed them into Stochastic Product Form.

Note that the new weights $W^\n(\pivec) = \frac{1}{Z} \What(\pivec)$
sum to 1 (i.e. Property~(3) holds) since $Z$ normalizes the weights.
Our goal is to find new multiedge weights $w_m^\n$
so that the other two properties hold as well, i.e.~
$W^\n(\pivec) = \prod_{m \in M} (w_{m}^\n)^{\pi_{m}}$ for all multipaths $\pivec$
and $\sum_{m\in M^\text{(out)}_v} w_m^\n=1$ for all nonsinks $v$.
For this purpose, we introduce a normalization $Z_v$ for each vertex $v \in V$:
\begin{equation}
Z_v := \sum_{\pivec \in \Pcal_v} \What(\pivec) .
\label{eq:z_v_def}
\end{equation}
where $\Pcal_v$ is the set of all multipaths sourced from $v$ and sinking at $\Tcal$.
Intuitively, $Z_v$ is the normalization constant for the subgraph sourced at $v \in V$ and sinking in $\Tcal$.
Thus for a sink node $v \in \Tcal$, $Z_v = 1$.
Moreover $Z = Z_s$ is the normalization factor for the multi-DAG $\Hcal$ 
where $s \in V$ is the source node.
The \textit{generalized weight pushing} finds all $Z_v$'s recursively starting from the sinks
and then it computes the new weights $w_m^\n$ for the multiedges to be used in the next trial:

\begin{enumerate}
\item For sinks $v\in \Tcal$, $Z_v = 1$.
\item Recursing backwards in the multi-DAG, 
let $Z_v\! =\! \sum_{m \in M^\text{(out)}_v}  \what_{m} \!\prod_{u \in U: m=(v,U)} Z_u$
for all non-sinks $v$.
\item For each multiedge $m=(v,U)$, $w_{m}^\n := \what_{m} \, \big(\prod_{u \in U} Z_{u} \big) / {Z_v}$.
\end{enumerate}

Figure~\ref{fig:weight_pushing_ex} illustrates an example of the weight pushing algorithm.
For simplicity, we demonstrate this algorithm on a regular DAG,
that is, a multi-DAG where $|U|=1$ for all multiedges $(v,U) \in M$.
The DAG on the left shows the unnormalized weights $\what_m$ for all multiedges/edges $m$ in the DAG.
In the DAG in the middle,  we compute all the \Green{normalizations} $\Green{Z_v}$ for all vertices $v \in V$
using Steps~1 and 2 of the weight pushing algorithm.
Finally, in the DAG on the right, we find the \Purple{new weights} $\Purple{w^\n_m}$ which are in Stochastic Product Form
using Step~3 of the weight pushing algorithm.
Lemma below proves the correctness and time complexity of this generalized weight pushing algorithm.

\begin{figure}[t]
\begin{minipage}[R]{0.3\textwidth}
\centering
\scalebox{.9}{
\begin{tikzpicture}%
  [>=stealth,
   shorten >=1pt,
   node distance=2cm,
   on grid,
   auto,
   every state/.style={draw=black!60, fill=black!5}
  ]
  
\tikzset{cutting/.style = {->,> = latex, very thick, yellow}}
    \tikzset{rectangle/.style={draw=gray,dashed,fill=green!1,thick,inner sep=5pt}}

\node[text width=1cm] at (-1,-2.2) {\large $s$};
\node[text width=1cm] at (1.7,2.2) {\large $\Tcal$};
  
\node[state] (2) { } ;
\node[state, very thick] (0) [below left=of 2] { };
\node[state] (1) [below right=of 2] { };
\node[state] (3) [above left=of 2] { };
  \node[state, accepting] (4) [above right=of 2] { };

\path[->]
	(0) edge	node {$3$} (1)
		  edge   node {$2$} (2)
		  edge   node {$1$} (3)
	(1) edge	node {$1$} (4)
	(2) edge	node {$2$} (3)
	      edge	node {$3$} (4)
	(3) edge	node {$1$} (4)
   ;
   

\end{tikzpicture}
}
\end{minipage}
\begin{minipage}[R]{0.3\textwidth}
\centering
\scalebox{.9}{
\begin{tikzpicture}%
  [>=stealth,
   shorten >=1pt,
   node distance=2cm,
   on grid,
   auto,
   every state/.style={draw=black!60, fill=black!5 }
  ]
  
\tikzset{cutting/.style = {->,> = latex, very thick, yellow}}
    \tikzset{rectangle/.style={draw=gray,dashed,fill=green!1,thick,inner sep=5pt}}

\node[text width=1cm] at (-1,-2.2) {\large $s$};
\node[text width=1cm] at (1.7,2.2) {\large $\Tcal$};

\node[state] (2) {\Green{\textbf{5}}} ;
\node[state, very thick] (0) [below left=of 2] {\Green{\textbf{14}}};
\node[state] (1) [below right=of 2] {\Green{\textbf{1}}};
\node[state] (3) [above left=of 2] {\Green{\textbf{1}}};
  \node[state, accepting] (4) [above right=of 2] {\Green{\textbf{1}}};

\path[->]
	(0) edge	node {$3$} (1)
		  edge   node {$2$} (2)
		  edge   node {$1$} (3)
	(1) edge	node {$1$} (4)
	(2) edge	node {$2$} (3)
	      edge	node {$3$} (4)
	(3) edge	node {$1$} (4)
   ;
   

\end{tikzpicture}
}
\end{minipage}
\begin{minipage}[R]{0.3\textwidth}
\hspace{1cm}
\centering
\scalebox{.9}{
\begin{tikzpicture}%
  [>=stealth,
   shorten >=1pt,
   node distance=2cm,
   on grid,
   auto,
   every state/.style={draw=black!60, fill=black!5 }
  ]
  
\tikzset{cutting/.style = {->,> = latex, very thick, yellow}}
    \tikzset{rectangle/.style={draw=gray,dashed,fill=green!1,thick,inner sep=5pt}}

\node[text width=1cm] at (-1,-2.2) {\large $s$};
\node[text width=1cm] at (1.7,2.2) {\large $\Tcal$};

\node[state] (2) {\Green{\textbf{5}}} ;
\node[state, very thick] (0) [below left=of 2] {\Green{\textbf{14}}};
\node[state] (1) [below right=of 2] {\Green{\textbf{1}}};
\node[state] (3) [above left=of 2] {\Green{\textbf{1}}};
  \node[state, accepting] (4) [above right=of 2] {\Green{\textbf{1}}};

\path[->]
	(0) edge	node {$\Purple{\mathbf{\frac{3}{14}}}$} (1)
		  edge   node {$\Purple{\mathbf{\frac{10}{14}}}$} (2)
		  edge   node {$\Purple{\mathbf{\frac{1}{14}}}$} (3)
	(1) edge	node {$\Purple{\mathbf{1}}$} (4)
	(2) edge	node {$\Purple{\mathbf{\frac{2}{5}}}$} (3)
	      edge	node {$\Purple{\mathbf{\frac{3}{5}}}$} (4)
	(3) edge	node {$\Purple{\mathbf{1}}$} (4)
   ;
   

\end{tikzpicture}
}
\end{minipage}
\caption[Example of weight pushing for regular DAGs]
{Example of weight pushing for regular DAGs
i.e.\ when $|U|\!=\!1$ for all multiedges $(v,U)\!\in\!M$.
(Left) the unnormalized weights $\what_m$ for all multiedges/edges $m$ in the DAG.
(Middle) the \Green{normalizations} $\Green{Z_v}$ for all vertices $v\!\in\!V$
using the Steps~1 and 2 of the weight pushing algorithm.
(Right) the \Purple{new weights} $\Purple{w^\n_m}$ which are in Stochastic Product Form
using the Step~3 of the weight pushing algorithm.
}
\label{fig:weight_pushing_ex}
\end{figure}

\begin{lemma}
The weights $w_{m}^\n$ generated by the generalized weight pushing are in Stochastic Product Form (see  Definition \ref{def:stochastic})
and for all multipaths $\pivec$, $\prod_{m \in M} (w_{m}^\n)^{\pi_{m}} = \frac{1}{Z_s} \prod_{m \in M} (\what_{m})^{\pi_{m}}$.
Moreover, the weights $w_{m}^\n$ can be computed in $\Ocal( c \, |M| )$ time
where $c$ is an upper-bound on the branching factor of each multiedge (i.e.~for all $m = (v, U) \in M$, $|U| < c$).
\end{lemma}

\begin{newproof}
First, we show that the recursive relation in Step~2 and the initialization in Step~1 hold for 
$Z_v$ defined in Equation (\ref{eq:z_v_def}).
For a sink node $v \in \Tcal$, the normalization constant $Z_v$ is vacuously $1$.
Thus Step~1 is justified.
To prove the recursive relation in Step 2, consider any non-sink $v \in V - \Tcal$.
We ``peel off'' the first multiedge leaving $v$ and then recurse:
$$
Z_v = \sum_{\pivec \in \Pcal_v} \What(\pivec) 
=\sum_{m \in M^\text{(out)}_v} \sum_{\substack{\pivec \in \Pcal_v \\
\text{ starts with } m
}}  \What(\pivec).
$$

Recall that $\What(\pivec) = \prod_{m' \in M} (\what_{m'})^{\pi_{m'}}$.
Thus, we can factor out the weight $\what_m$ associated with multiedge $m \in M^\text{(out)}_v$.  
Assume the multiedge $m$ comprised of edges from the node $v$ to the nodes $u_1, \ldots, u_k$.
Notice, excluding $m$ from the multipath, we are left with $k$ number of multipaths from the $u_i$'s:
\begin{align*}
Z_v &=\sum_{m \in M^\text{(out)}_v} \sum_{\substack{\pivec \in \Pcal_v \\
\text{ starts with } m
}}  \What(\pivec)  \\
&=\sum_{m \in M^\text{(out)}_v} \what_{m}
\sum_{\pivec_1 \in \Pcal_{u_1}}
\sum_{\pivec_2 \in \Pcal_{u_2}}
\cdots
\sum_{\pivec_k \in \Pcal_{u_k}}
\prod_{i=1}^k \What(\pivec_{i}).
\end{align*}

After factoring $\what_m$ out, the sum 
$\sum_{\pivec_1 \in \Pcal_{u_1}}
\sum_{\pivec_2 \in \Pcal_{u_2}}
\cdots
\sum_{\pivec_k \in \Pcal_{u_k}}$
iterates over all combinations of all multipaths sourced from all $u_i$'s associated with the multiedge $m$. 
Recall that $\Pcal_{u_i}$ is the set of all multipaths sourced from $u_i$ and sinking at $\Tcal$.
Since each $\pivec_i$ iterates over all multipaths in $\Pcal_{u_i}$,
we can turn the sum of products into product of sums as below:
\begin{align}
Z_v &=\sum_{m \in M^\text{(out)}_v} \what_{m}
\sum_{\pivec_1 \in \Pcal_{u_1}}
\sum_{\pivec_2 \in \Pcal_{u_2}}
\cdots
\sum_{\pivec_k \in \Pcal_{u_k}}
\prod_{i=1}^k \What(\pivec_{i}) \nonumber \\
&= \sum_{m \in M^\text{(out)}_v} \what_{m}
\sum_{\pivec_1 \in \Pcal_{u_1}}
\What(\pivec_{1}) 
\underbrace{\sum_{\pivec_2 \in \Pcal_{u_2}}
\cdots
\sum_{\pivec_k \in \Pcal_{u_k}}
\prod_{i=2}^k \What(\pivec_{i}) )}_{\text{does not depend on } \pivec_1 \in \Pcal_{u_1}}
\nonumber \\
&= \sum_{m \in M^\text{(out)}_v} \what_{m}
\left(
\sum_{\pivec_2 \in \Pcal_{u_2}}
\cdots
\sum_{\pivec_k \in \Pcal_{u_k}}
\prod_{i=2}^k \What(\pivec_{i}) 
\right)
\left(
\sum_{\pivec_1 \in \Pcal_{u_1}}
\What(\pivec_{1}) 
\right) \nonumber\\
&= \qquad \cdots \nonumber \qquad \qquad \text{(Repeating for each sum $\sum_{\pivec_j \in \Pcal_{u_j}}$ )}\\
&=\sum_{m \in M^\text{(out)}_v} \what_{m}
\prod_{i=1}^k 
\left(
\underbrace{\sum_{ \pivec \in \Pcal_{u_i}}  \What(\pivec) }_{Z_{u_i}} 
\right)
\nonumber \\
&=\sum_{m \in M^\text{(out)}_v} \what_{m}
\prod_{i = 1}^k {Z_{u_i}}. \label{eq:z_v}
\end{align}

Equation~(\ref{eq:z_v}) justifies Step~2.
Now we prove that the new weight assignment in Step~3 
will result in a distribution in Stochastic Product Form with correct expectation.
For all $v \in V - \Tcal$ and for all $m \in M^\text{(out)}_v$, set $w_{m}^\n := \what_{m} \, \frac{\prod_{u: (v,u) \in m} Z_u}{Z_v}$ (Step~3).
Property~(2) of Definition~\ref{def:stochastic} (i.e. stochasticity) is true since for all $v \in V - \Tcal$:
\begin{align*}
\sum_{m \in M^\text{(out)}_v} w^\n_{m} 
&= \sum_{m \in M^\text{(out)}_v} \what_{m} \, \frac{\prod_{u: (v,u) \in m} Z_u}{Z_v} \\
&= \frac{1}{Z_v} \, \underbrace{\sum_{m \in M^\text{(out)}_v} \what_{m} \, \prod_{u: (v,u) \in m} Z_u}_{Z_v}=1. && \text{(Equation (\ref{eq:z_v}))}
\end{align*}

We now prove that Property~(1) of Definition~\ref{def:stochastic} (i.e. product form) is also true since for all $\pivec \in \Pcal_s$:
\begin{align*}
\prod_{m \in M} (w^\n_{m})^{\pi_{m}}
&= \prod_{v \in V-\Tcal} \prod_{m \in M^\text{(out)}_v} (w^\n_{m})^{\pi_{m}} \\
&= \prod_{v \in V-\Tcal} \prod_{m \in M^\text{(out)}_v} \rbr{\what_{m} \, \frac{\prod_{u: (v,u) \in m} Z_u}{Z_v}}^{\pi_{m}} \\
&= \sbr{\prod_{v \in V-\Tcal} \prod_{m \in M^\text{(out)}_v} \rbr{\what_{m} }^{\pi_{m}}} \, 
\sbr{\prod_{v \in V-\Tcal} \prod_{m \in M^\text{(out)}_v} \rbr{\frac{\prod_{u: (v,u) \in m} Z_u}{Z_v}}^{\pi_{m}}}.
\end{align*}

Notice that $\prod_{v \in V-\Tcal} \prod_{m \in M^\text{(out)}_v} \rbr{\frac{\prod_{u: (v,u) \in m} Z_u}{Z_v}}^{\pi_{m}}$ telescopes along the multiedges in the multipath $\pivec$. 
After telescoping, since $Z_v=1$ for all $v \in \Tcal$, the only remaining term will be $\frac{1}{Z_s}$ where $s$ is the souce node.
Therefore we obtain:
\begin{align*}
\prod_{m \in M} (w^\n_{m})^{\pi_{m}}
&= \sbr{\prod_{v \in V-\Tcal} \prod_{m \in M^\text{(out)}_v} \rbr{\what_{m} }^{\pi_{m}}} \, 
\sbr{\prod_{v \in V-\Tcal} \prod_{m \in M_v} \rbr{\frac{\prod_{u: (v,u) \in m} Z_u}{Z_v}}^{\pi_{m}}} \\
&= \sbr{ \prod_{m \in M} \rbr{\what_{m}}^{\pi_{m}}} \, 
\sbr{ \frac{ 1 }{Z_s} } \\
&= \frac{ 1}{Z_s} \, { \prod_{m \in M} \rbr{\what_{m}}^{\pi_{m}}} = W^\n(\pivec).
\end{align*}

Regarding the time complexity, we first focus on the the recurrence relation 
$Z_v = \sum_{m \in M_v}  \what_{m}\,  \prod_{u: (v,u) \in m} Z_u$.
Note that for each $v \in V$, $Z_v$ can be computed in $\Ocal( c \, |M_v^\text{(out)}|  )$.
Thus the computation of all $Z_v$'s takes $\Ocal( c \, |M|  )$ time. 
Now observe that  $w_{m}^\n$ for each multiedge $m = (v, U)\in M$ can be found in $\Ocal( c )$ time using $w_{m}^\n = \what_{m}\, \frac{\prod_{u \in U} Z_u}{Z_v}$.
Hence the computation of $w_{m}^\n$ for all multiedges $m \in M$ takes $\Ocal( c \, |M| )$ time.
Therefore the generalized weight pushing algorithm runs in $\Ocal( c \, |M| )$ time.

\end{newproof}

\paragraph{Regret Bound.}

In order to apply the regret bound of EH 
we have to initialize the distribution $W$ on multipaths to the uniform distribution. 
This is achieved by setting all the weights $\what_m$'s to $1$ followed by an application of generalized weight pushing.
Note that Theorem~\ref{thm:EH-paths} is a special case of the theorem below when $|U|=1$ for all multiedge $(v,U) \in M$ and $|\Tcal|=1$.

\begin{theorem} 
\label{thm:EH-multipaths}
Given a multi-DAG $\Hcal = (V,M)$ with designated source node $s \in V$ and sink nodes $\Tcal \subset V$, 
assume $\Ncal$ is the number of multipaths in $\Hcal$ from $s$ to $\Tcal$, 
$L^*$  is the total loss of best multipath,
and $D$ is an upper-bound on the $1$-norm of the count vectors of the multipaths (i.e.\ $\| \pivec \|_1 \leq D$ for all multipaths $\pivec$).
Then with proper tuning of the learning rate $\eta$ over the trials, EH guarantees the following regret bound:
$$
\mathcal{R}_\text{EH} \leq \sqrt{2 \, L^* \, D \, \log \Ncal} + D \, \log \Ncal.
$$
\end{theorem}

\subsection{Component Hedge on Multipaths}
\label{sec:ch-multipaths}

We implement CH efficiently for learning multipaths in a multi-DAG $\Hcal = (V, M)$.
The multipaths are represented as $|M|$-dimensional count vectors $\pivec$ (see Definition~\ref{def:multipath}).
The algorithm maintains an $|M|$-dimensional mixture vector $\fvec$ in the convex hull of count vectors. 
This hull is the following polytope obtained by relaxing the integer constraints on the count vectors:

\begin{definition}[Unit-Flow Polytope]
\label{def:unit-flow-polytope}
Given a multi-DAG $\Hcal = (V,M)$, let 
 $\fvec \in \RR_{\geq 0}^{|M|}$ in which $f_m$ is associated with $m \in M$. 
 Define the inflow $f_\text{in}(v) := \sum_{m \in M^\text{(in)}_v} f_m$ 
and the outflow $f_\text{out}(v) := \sum_{m \in M^\text{(out)}_v} f_m$.
$\fvec$ belongs to the \textbf{unit-flow polytope} of $\Hcal$ if it has the following
properties:
\begin{enumerate}
\item The outflow $f_\text{out}(s)$ of the source $s$ is $1$.
\item For each vertex $v \in V\!-\!\Tcal\!-\!\{s\}$, 
the outflow is equal to the inflow, i.e. $f_\text{out}(v) = f_\text{in}(v)$.
\end{enumerate}
\end{definition}

In each trial, the weight of each multiedge $f_m$ is updated multiplicatively to $\fhat_m = f_m \exp(-\eta\ell_m)$ 
and then the weight vector $\fvechat$ is projected back to the unit-flow polytope via a relative entropy projection:
$$
\fvec^\n := 
\underset{\fvec \in \text{unit-flow polytope}}{\arg\min} 
\Delta(\fvec ||
\fvechat), \quad \text{ where } \quad \Delta(\avec ||
\bvec) = \sum_i a_i \log \frac{a_i}{b_i} + b_i - a_i.
$$

This projection is achieved by repeatedly cycling over the vertices
and project onto the local flow constraints at the current vertex. 
This method is called iterative Bregman projections \citep{bregman1967relaxation}. 
The following lemma shows that projection to each local flow constraint
is simply equivalent to scaling the in- and out-flows to the appropriate values.

\begin{lemma}
The relative entropy projection to the local flow constraint at vertex $v \in V$ is done as follows:
\begin{enumerate}
\item If $v=s$, normalize the $f_\text{out}(v)$ to $1$.
\item If $v \in V\! -\! \Tcal\! -\! \{s\}$, 
scale the incoming and outgoing multiedges of $v$ such that
$$f_\text{out}(v) := f_\text{in}(v) := \sqrt{ f_\text{out}(v) \cdot f_\text{in}(v)}.$$
\end{enumerate}
\end{lemma}

\begin{newproof}
Formally, the projection $\fvec$ of a given point $\fvechat \in \RR_{\geq 0}^{|M|}$ to constraint $C$ is the solution to the following:
\begin{displaymath}
\underset{\fvec \in C}{\arg\min} \sum_{m \in M} \, f_m \log \rbr{\frac{f_m}{\fhat_m}} + \fhat_m - f_m.
\end{displaymath}

$C$ can be one of the two types of constraints mentioned in Definition \ref{def:unit-flow-polytope}. 
We use the method of Lagrange multipliers in both cases. 
Observe that if $|U|=1$ for all multiedge $m=(v,U) \in M$, then the updates in \cite{koolen2010hedging} are recovered.

\paragraph{Constraint Type 1.} 
The outflow from the source $s$ must be $1$. Assume $f_{m_1}, \ldots, f_{m_d}$ are the weights 
associated with the outgoing multiedges $m_1, \ldots, m_d$ from the source $s$. 
Then:
\begin{align}
&L(\fvec, \lambda) := \sum_{m \in M} \, f_m \log \rbr{\frac{f_m}{\fhat_m}} + \fhat_m - f_m - \lambda \left( \sum_{j=1}^d f_{m_j} - 1 \right) \nonumber \\
& \frac{\partial L}{\partial f_m} = \log \frac{f_m}{\fhat_m} = 0 \, \longrightarrow \, f_m = \fhat_m  \qquad \forall m \in M - \{m_1, \ldots, m_d \} \nonumber \\
& \frac{\partial L}{\partial f_{m_j}} = \log \frac{f_{m_j}}{\fhat_{m_j}} - \lambda = 0 \, \longrightarrow \, f_{m_j}= \fhat_{m_j} \, \exp(\lambda) \label{eq:root_norm1}\\
& \frac{\partial L}{\partial \lambda} = \sum_{j=1}^d f_{m_j} - 1 = 0. \label{eq:root_norm2}
\end{align}

Combining equations (\ref{eq:root_norm1}) and (\ref{eq:root_norm2}) results in normalizing $f_{m_1}, \ldots, f_{m_d}$, that is:
\begin{displaymath}
\forall j \in \{1..d\} \quad f_{m_j}= \frac{ \fhat_{m_j}}{\sum_{j'=1}^d \fhat_{m_{j'}}}.
\end{displaymath}

\paragraph{Constraint Type 2.} 
Given any internal node $v \in V - \Tcal - \{s\}$, the outflow from $v$ must be equal to the inflow of $v$. 
Assume $f^{(\text{in})}_{1}, \ldots, f^{(\text{in})}_{a}$ and $f^{(\text{out})}_{1}, \ldots, f^{(\text{out})}_{b}$ are the weights associated with 
the incoming and outgoing multiedges from/to the node $v$, respectively. Then:
\begin{align}
&L(\wvec, \lambda) := \sum_{m \in M} \, f \log \rbr{\frac{f_m}{\fhat_m}} + \fhat_m - f_m - \lambda \left( \sum_{b'=1}^b f^{(\text{out})}_{b'} -  \sum_{a'=1}^a f^{(\text{in})}_{a'} \right)  \nonumber \\
& \frac{\partial L}{\partial f_m} = \log \frac{f_m}{\fhat_m} = 0 \, \longrightarrow \, f_m = \fhat_m  \quad \forall m \text{ non-adjacent to } v \nonumber \\
& \frac{\partial L}{\partial f^{(\text{out})}_{{b'}}} = \log \frac{f^{(\text{out})}_{{b'}}}{\fhat^{(\text{out})}_{{b'}}} - \lambda = 0 \, \longrightarrow \, f^{(\text{out})}_{{b'}}= \fhat^{(\text{out})}_{{b'}} \, \exp(\lambda) \qquad \forall b' \in \{1..b\} \label{eq:vScaling1}\\
& \frac{\partial L}{\partial f^{(\text{in})}_{{a'}}} = \log \frac{f^{(\text{in})}_{{a'}}}{\fhat^{(\text{in})}_{{a'}}} +  \lambda = 0 \, \longrightarrow \, f^{(\text{in})}_{{a'}}= \fhat^{(\text{in})}_{{a'}} \, \exp(- \lambda) \qquad \forall a' \in \{1..a\} \label{eq:vScaling2}\\
& \frac{\partial L}{\partial \lambda} = \sum_{b'=1}^b f^{(\text{out})}_{{b'}} -  \sum_{a'=1}^a f^{(\text{in})}_{{a'}} = 0. \label{eq:vScaling3}
\end{align}

Letting $\beta = \exp(\lambda)$, for all $a' \in \{1..a\}$ and all $b' \in \{1..b\}$, we can obtain the following by combining equations (\ref{eq:vScaling1}), (\ref{eq:vScaling2}) and (\ref{eq:vScaling3}):
$$
\beta \, \left( \sum_{b'=1}^b \fhat^{(\text{out})}_{{b'}} \right) = \frac{1}{\beta} \left( \sum_{a'=1}^a \fhat^{(\text{in})}_{{a'}} \right)
\longrightarrow \beta = \sqrt{\frac{\sum_{a'=1}^a \fhat^{(\text{in})}_{{a'}}}{\sum_{b'=1}^b \fhat^{(\text{out})}_{{b'}}}}
$$
$$
\forall \, b' \in \{1..b\}, \;
f^{(\text{out})}_{{b'}}= \fhat^{(\text{out})}_{{b'}} \, \sqrt{\frac{\sum_{a''=1}^a \fhat^{(\text{in})}_{{a''}}}{\sum_{b''=1}^b \fhat^{(\text{out})}_{{b''}}}},
$$
$$
\forall \, a' \in \{1..a\}, \;
f^{(\text{in})}_{{a'}}= \fhat^{(\text{in})}_{{a'}} \, \sqrt{\frac{\sum_{b''=1}^b \what^{(\text{out})}_{{b''}}}{\sum_{a''=1}^a \what^{(\text{in})}_{{a''}}}}.
$$

This indicates that to enforce the flow conservation property at each internal node, the weights must be multiplicatively scaled up/down 
so that the new outflow and inflow is the geometric average of the old outflow and inflow. 
\end{newproof}

\paragraph{Prediction.} 
In this step, the algorithm needs to randomly predict with a multipath $\pivec$ from a distribution $\Dcal$ such that $\EE_\Dcal [ \pivec] = \fvec$.
In Component Hedge and similar algorithms \citep{helmbold2009learning, koolen2010hedging, yasutake2011online, warmuth2008randomized},
$\Dcal$ is constructed by decomposing $\fvec$ into a convex combination of small number of solutions.
In this section, we give a new and more direct prediction method for multipaths. 
We construct a distribution $\Dcal$ with the right expectation 
in Stochastic Product Form (see Definition~\ref{def:stochastic}) 
by defining a new set of weights $\wvec$ using the flow values $\fvec$.
For each multiedge $m=(v, U) \in M$, we set the weight $w_m = f_m / f_\text{in}(v)$.
The induced distribution will be in Stochastic Product Form with the right expectation $\EE_\Dcal [ \pivec] = \fvec$.
This gives us a faster prediction method as the decomposition is avoided.
Lemma~\ref{lem:ch2eh} shows the correctness and time complexity of our method.

\begin{lemma}
\label{lem:ch2eh}
For each multiedge $m=(v, U) \in M$, define the weight $w_m = f_m / f_\text{in}(v)$. 
Let the distribution $\Dcal$ over the multipaths be $\Dcal(\pivec) := \prod_{m \in M} (w_{m})^{\pi_{m}}$.
Then:
\begin{enumerate}
\item $\Dcal$ is in Stochastic Product Form.
\item $\EE_\Dcal [ \pivec] = \fvec$.
\item Constructing $\Dcal$ from the flow values $\fvec$ can be done in $\Ocal(c |M|)$ time
where $c$ is an upper-bound on the branching factor of each multiedge (i.e. for all $m = (v,U) \in M$, $|U| < c$).
\end{enumerate}
\end{lemma}

\begin{newproof}
$\Dcal(\pivec)$ is in product form by construction. 
The weights are also stochastic since for each non-sink vertex $v$:
$$
\sum_{m \in M^\text{(out)}_v} w_m 
= \sum_{m \in M^\text{(out)}_v} \frac{f_m }{f_\text{in}(v)}
= \frac{1}{f_\text{in}(v)} \sum_{m \in M^\text{(out)}_v} f_m 
= \frac{1}{f_\text{in}(v)} f_\text{out}(v) = 1
$$

Thus the $\Dcal$ is in Stochastic Product Form (Definition~\ref{def:stochastic}). 
Now we show that $\Dcal$ will result in the desired expectation. 
Let $\fvechat := \EE_\Dcal [ \pivec] $ be the flow induce by $\Dcal$. 
Denote $\fhat_\text{in}(v) := \sum_{m \in M^\text{(in)}_v} \fhat_m$.
Let $v_1, \ldots, v_n$ be a topological order of the vertices in the underlying DAG.
We use strong induction on $n$ to show that $\fhat_\text{in}(v) = f_\text{in}(v)$ for all $v \in V$.
For $v_1=s$ this is true since $\fhat_\text{in}(s) = f_\text{in}(s)=1$. 
For $i > 1$:
\begin{align*}
\fhat_\text{in}(v_i) 
&= \sum_{m = (v, U) \in M^\text{(in)}_{v_i}} w_m \, \fhat_\text{in}(v) \\
&= \sum_{m = (v, U) \in M^\text{(in)}_{v_i}} w_m \, f_\text{in}(v) && \text{(Inductive hypothesis)}\\
&= \sum_{m = (v, U) \in M^\text{(in)}_{v_i}} \frac{f_m}{f_\text{in}(v)} \, f_\text{in}(v) && \text{(Definition of $w_m$)} \\
&= \sum_{m = (v, U) \in M^\text{(in)}_{v_i}} f_m = f_\text{in}(v_i) 
\end{align*}
and that completes the induction. Now for each multiedge $m = (v,U)\in M$ we have:
$$
\fhat_m = \fhat_\text{in}(v)  w_m = \fhat_\text{in}(v)  \frac{f_m}{f_\text{in}(v)} = f_m
$$
Thus $\fvec = \fvechat$. 

To construct $\Dcal$, we must find all the weights $w_m$. 
To do so, we will have two passes over the set of multiedges $M$.
In the first pass, we compute all incoming flows $f_\text{in}(v)$ for all $v \in V$  in $O(c |M|)$ time.
Then in the second pass we find all the weights $w_m =\frac{f_m}{f_\text{in}(v)}$  in $O(|M|)$ time.
Having constructed $\Dcal$, we can efficiently sample a multipath with the right expectation.
\end{newproof}

\paragraph{Regret Bound.} 
The regret bound for CH depends on
a good choice of the initial weight vector $\fvec^\text{init}$ 
in the unit-flow polytope.
We use an initialization technique 
similar to the one discussed in \cite{rahmanian2018xfhedge}.
Instead of explicitly selecting $\fvec^\text{init}$ in the unit-flow polytope, 
the initial weight is obtained by projecting a point $\fvechat^{\text{init}}$
outside of the polytope to its inside. This yields the
following regret bounds.
\begin{theorem}
\label{thm:CH-multipaths}
Given a multi-DAG $\Hcal = (V,M)$,
let $D$ be an upper-bound on the $1$-norm of the count vectors of the multipaths (i.e.\ $\| \pivec \|_1 \leq D$ for all multipaths $\pivec$).
Also denote the total loss of the best multipath by $L^*$. 
Then with proper tuning of the learning rate $\eta$ over the trials, CH guarantees:
$$
\mathcal{R}_\text{CH} \leq \sqrt{2 \, L^* \, D \, (\log |M| +  \log D) } + D \, \log |M| + D \log D.
$$

Moreover, when the multipaths are bit vectors, then:
$$
\mathcal{R}_\text{CH} \leq \sqrt{2 \, L^* \, D \, \log |M|} + D \log |M|.
$$
\end{theorem}
\begin{newproof}
According to \cite{koolen2010hedging}, with proper tuning of the learning rate $\eta$, the regret bound of CH is:
\begin{equation}\label{eq:CH-general-regret}
\mathcal{R}_\text{CH}  \leq \sqrt{2 \, L^* \, \Delta(\pivec || \fvec^\text{init}) } +  \Delta(\pivec || \fvec^\text{init}),
\end{equation}
where $\pivec \in \NN^{|M|}$ is the best multipath and $L^*$ its loss.
Define $\fvechat^\text{init} := \frac{1}{|M|} \, \one$ where $\one \in \RR^{|M|}$ is a vector of all ones. 
Now let the initial point $\fvec^\text{init}$ be the relative entropy projection of $\fvechat^\text{init}$ onto the unit-flow prolytope\footnote{This computation can be done as a pre-processing step.}
$$\fvec^\text{init} = \arg\min_{\fvec \in P} \Delta(\fvec || \fvechat^\text{init}).$$

Now we have:
\begin{align}
\Delta(\pivec || \fvec^\text{init}) 
&\leq \Delta(\pivec || \fvechat^\text{init}) && \text{\hspace{-3cm} (Generalized Pythagorean Thm.)} \nonumber \\
&= \sum_{m \in M} \pi_m \log \frac{\pi_m}{\fhat^\text{init}_m} +\fhat^\text{init}_m - \pi_m \nonumber \\
&= \sum_{m \in M} \pi_m \log \frac{1}{\fhat^\text{init}_m} + \pi_m \log \pi_m +\fhat^\text{init}_m - \pi_m \nonumber \\
&\leq \sum_{m \in M} \pi_m (\log |M|) + \sum_{m \in M} \pi_m \log \pi_m + \sum_{m \in M} \frac{1}{|M|} -  \sum_{m \in M} \pi_m \label{eq:ch-bound-cont} \\
&\leq D (\log |M|) + D \log D + |M| \, \frac{1}{|M|} -  \sum_{m \in M} \pi_m \nonumber \\
&\leq  D \, \log |M| + D \log D. \nonumber 
\end{align}

Thus, by Inequality (\ref{eq:CH-general-regret}) the regret bound will be:
$$\mathcal{R}_\text{CH} \leq \sqrt{2 \, L^* \, D \, (\log |M| + \log D)  } 
+  D  \, \log |M| + D \log D. $$

Note that if $\pivec$ is a bit vector, then $\sum_{m \in M} \pi_m \log \pi_m = 0$, and consequently, the expression (\ref{eq:ch-bound-cont}) can be bounded as follows:
\begin{align*}
\Delta(\pivec || \fvec^\text{init}) 
&\leq \sum_{m \in M} \pi_m (\log |M|) + \sum_{m \in M} \pi_m \log \pi_m + \sum_{m \in M} \frac{1}{|M|} -  \sum_{m \in M} \pi_m  \\
&\leq D (\log |M|) + |M| \, \frac{1}{|M|} -  \sum_{m \in M} \pi_m \nonumber \\
&\leq D \, \log |M|. \nonumber 
\end{align*}

Again, using Inequality (\ref{eq:CH-general-regret}), the regret bound will be:
$$\mathcal{R}_\text{CH}  \leq \sqrt{2 \, L^* \, D \, \log |M|   } 
+  D  \, \log |M|. $$
\end{newproof}

Notice that by setting $|U|=1$ for all multiedge $(v,U) \in M$ and $|\Tcal|=1$, the algorithm for path learning in \cite{koolen2010hedging} is recovered. 
Also observe that Theorem~\ref{thm:CH-paths} is a corollary of Theorem~\ref{thm:CH-multipaths} since every path is represented as a bit vector
and $|M| = |E| \leq |V|^2$.

\subsection{Stochastic Product Form vs Mean Form}
\label{sec:odp:mapping}

We discussed the efficient implementation of the two algorithms of EH and CH for learning multipaths.
The EH algorithm maintains a \emph{weight} vector $\wvec \in \RR^{|M|}$ in the Stochastic Product Form.
These weights define a distribution over all multipaths.
On the other hand, the CH algorithm keeps track of a \emph{flow} vector $\fvec \in \RR^{|M|}$ in the \emph{Mean Form}.
These flows define a mean vector over all multipaths and belong to the unit-flow polytope.

For any distribution over the multipaths, 
there is a unique expectation/mean of the counts of the multiedges according to the given distribution.
This expectation is represented by a flow vector.
If the distribution is in Stochastic Product Form with the weight vector $\wvec$, 
the flow vector can be computed efficiently using a dynamic programming algorithm.
Initializing with the source $s$, we set the in-coming flow $f_{\text{in}}(s) = 1$.
Then, using the recursive equation $f_m = w_m f_\text{in}(s)$ for all $m \in M_s^\text{(out)}$,
we find the flows of the out-going multiedges from the source $s$ by partitioning the in-flow according to its out-going weights.
Having computed the flows of all the out-going multiedges, we can find the in-flows of some of the vertices which are connected to the source. 
By applying the aforementioned recursion over the vertices of $\Hcal$ in the topological order of the underlying base directed graph $\Bcal(\Hcal)$,
we can find the flows of all the multiedges.
This procedure can be done in $\Ocal(c |M |)$ time where
$c$ is an upper-bound on the branching factor of each multiedge (i.e. for all $m = (v,U) \in M$, $|U| < c$).

Conversely, by applying the Lemma~\ref{lem:ch2eh} on a given flow vector $\fvec$, 
we can find the weights $\wvec$ defining the distribution $\Dcal$ 
in the Stochastic Product Form such that it has the right expectation $\EE_\Dcal [\pivec] = \fvec$.
In general if we assume no structure on the distributions over the multipaths, 
there could be several different distributions with the expectation $\fvec$.
However, if we limit the distributions to the Stochastic Product Form, 
then the resulting distribution $\Dcal$  is unique.
This is because the in-flows should be distributed according to the local weights
in the Stochastic Product Form. 

\begin{figure}
\scalebox{1.0}{

\tikzset{
	bigArrow/.style={
		decoration={markings,mark=at position 1 with {\arrow[scale=3]{>}}},
		postaction={decorate},
		shorten >=0.4pt}
	}

\tikzstyle{space}=[draw, circle, very thick, black, text width = 0.8, align=center, fill=blue!5]

{ 
\begin{tikzpicture}

\node[space, label={\Large {$\wvec \in \RR^{|M|}$}}, %
	 label={[text width = 1.5in, align=left, gray]below:{\small •Stochastic \\•Multiplicative Updates \\•Weight Pushing }} ] %
	(orig) [text width = 0.9in, align=center]  {\Large Stochastic Product Form} ;

\node[space, label={\Large {$\fvec \in \RR^{|M|}$} }, %
	label={[text width = 1.9in, align=left, gray]below:{\small $\qquad$•Unit-Flow Polytope \\$\qquad$•Multiplicative Updates \\$\qquad$•Projection}  } ] %
	(ext) [right = 2.25in of orig, text width = 0.9in, align=center] {\Large Mean Form} ;

\tikzset{rectangle/.style={draw=purple,dashed,thick,inner sep=5pt}}

\node[rectangle, anchor=south west, minimum width=4cm, minimum height=6.7cm] (A) at (-2.1,-4.2) {};
\node[rectangle, anchor=south west, minimum width=4cm, minimum height=6.7cm] (A) at (6.8,-4.2) {};

\node[text width=5cm] at (1,3) {\large \Violet{Expanded Hedge}};
\node[text width=5cm] at (9.5,3) {\large \Violet{Component Hedge}};

\node[text width=5cm] at (5,0) {\large \Green{Preserving Mean}};

\draw[bigArrow, bend left, black, -latex, thick] (orig) to node[auto, black, text width = 2.0in, align=center] 
			{Dynamic Programming starting from $s\in V$ \\ \Blue{$\;f_m\!:=\!w_m \, f_\text{in}(v), \, m\!\in\!M^\text{(out)}_v$} } (ext);
			
\draw[bigArrow, bend left, black, -latex, thick] (ext) to node[auto, black, text width = 2.0in, align=center] 
			{``Conditional Outflow'' \\ in parallel \\ \Blue{$w_m\!:=\!\frac{f_m}{f_\text{in}(v)}, \; m\!\in\!M^\text{(out)}_v$}} (orig);

\end{tikzpicture}

}}
\caption{Mapping between Stochastic Product Form in EH and Mean Form in CH.}
\label{fig:map-ch-eh}
\end{figure}

This establishes a $1$-$1$ and onto mapping between the Stochastic Product Form of EH and the Mean Form of CH
(see Figure~\ref{fig:map-ch-eh}).
Both directions of the mapping have the additional crucial property of preserving the mean.

\section{Online Dynamic Programming with Multipaths} 
\label{sec:dp-games}

We consider the combinatorial online learning problems whose offline optimization problem can be solved efficiently 
via a dynamic programming algorithm defined by an arbitrary min-sum recurrence.
This is equivalent to repeatedly solving a variant of the same min-sum dynamic programming problem in successive trials.

We will use our definition of multi-DAG (see Definition~\ref{def:multidag}) to describe a representation of the dynamic programming problem. 
The vertex set $V$ is a set of subproblems to be solved. 
The source node $s \in V$ is the ``complete subproblem'' (i.e.~the original problem).
The sink nodes $\Tcal\subset V$ are the base subproblems. 
A multiedge from a node $v \in V$ to a set of nodes $U \subset V$ means that
a solution to the subproblem $v$ may use solutions to the (smaller) subproblems in $U$.
Denote the set of all multiedges by $M$.
A step of the dynamic programming recursion is thus represented by a multiedge. 
Denote the constructed directed multigraph by $\Hcal = (V,M)$.
A subproblem is never solved more than once in a dynamic programming.
Therefore base directed graph $\Bcal(\Hcal)$ is acyclic and $\Hcal$ is a multi-DAG.

There is a loss associated with any sink node in $\Tcal$.
Also with the recursions at the internal node $v$ a local loss will be added to the loss of the subproblems
that depends on $v$ and the chosen multiedge $m \in M_v^\text{(out)}$ leaving $v$.
We can handle arbitrary \Red{``min-sum'' } recurrences:
$$
\text{OPT}(v) = 
\begin{cases}
L_\Tcal(v) & v \in \Tcal \\
\Red{\min_{m \in M^\text{(out)}_v} \{ \sum_{u:(v,u)\in m} }
\text{OPT}(u) \Red{+} L_M(m) \Red{\}} & v \in V-\Tcal.
\end{cases}
$$

The problem of repeatedly solving an arbitrary min-sum dynamic programming problem over trials 
now becomes online learning of multipaths in $\Hcal$.
Note that due to the correctness of the dynamic programming, 
every possible solution to the dynamic programming can be encoded as a multipath in $\Hcal$ and vice versa. 

The loss of a given multipath is the sum of $L_M(m)$ over all multiedges $m$ in the multipath
plus the sum of $L_\Tcal(v)$ for all sink nodes $v$ at the bottom of the multipath.
To capture the same loss, we can alternatively define losses over the multiedges $M$. 
Concretely, for each multiedge $m=(v,U)$ define 
$\ell_m := L_M(m) + \sum_{u \in U} \mathbbm{1}_{\{u \in
\Tcal \}} L_\Tcal(u)$ 
where $\mathbbm{1}_{\{\cdot\}} $ is the indicator function%
\footnote{
The alternative losses over the multiedges may not be in $[0,1]$.
However, it is straight-forward to see if $\ell_m \in [0,b]$ for some $b$, the regret bounds for CH and EH will scale up accordingly.
}.

\section{Applications}
\label{sec:instantiation}

In this section, we apply our algorithms to various instances of online dynamic programming. 
In each instance, we define the problem, explore the dynamic programming representation and obtain the regret bounds.

\subsection{Binary Search Trees}
Recall again the online version of optimal binary
search tree (BST) problem \citep{thomas2001introduction}:
We are given a set of $n$ distinct keys $K_1 < K_2 < \ldots< K_n$.
In each trial, the algorithm predicts with a BST. 
Then the adversary reveals a probability vector $\pvec \in [0,1]^n$ with $\sum_{i=1}^n p_i = 1$.
For each $i$, $p_i$ indicates the \emph{search probability} for the key $K_i$. 
The loss is defined as the \emph{average search cost} in the predicted BST which is 
the average depth\footnote{Here the root starts at depth 1.} 
of all the nodes in the BST:
$$\text{loss}  = \sum_{i=1}^n \text{depth}(K_i) \cdot p_i .$$

\paragraph{Convex Hull of BSTs.}
Implementing CH requires a representation where 
not only the BST polytope has a polynomial number of facets, 
but also the loss must be linear over the components.
Since the average search cost is linear in the $\text{depth}(K_i)$ variables, 
it would be natural to choose these $n$ variables as the components for representing a BST. 
Unfortunately the convex hull of all BSTs when represented this
way is not known to be a polytope with a polynomial number of facets.
There is an alternate characterization of the convex hull of BSTs 
with $n$ internal nodes called the \textit{associahedron} \citep{loday2005multiple}. 
This polytope has polynomial in $n$ many facets but the average search cost is not linear in
the $n$ components associated with this polytope%
\footnote{Concretely, the $i$th component is $a_i \, b_i$ 
where $a_i$ and $b_i$ are the number of nodes in the left and right 
subtrees of the $i$th internal node $K_i$, respectively.}.
Thus CH cannot be applied to associahedron.

\paragraph{The Dynamic Programming Representation.}
The optimal BST problem can be solved via dynamic programming \citep{thomas2001introduction}.
Each subproblem is denoted by a pair $(i,j)$, for $1\leq i \leq n+1$ and $i-1\le j\le n$, 
indicating the optimal BST problem with the keys $K_i,\ldots,K_j$.
The base subproblems are $(i, i-1)$, for $1\le i\le n+1$ and the complete subproblem is $(1,n)$.
The BST dynamic programming problem uses the following \Red{min-sum} recurrence:
$$
\text{OPT}(i,j)\! =\! \begin{cases}
0 & \!j\!=\!i\!-\!1\\
\Red{\min_{i\leq r \leq j} \{ } \text{OPT}\Green{(i, r\!-\!1)} \Red{+} 
                         \text{OPT}\Green{(r\!+\!1, j)}
\Red{+}
\sum_{k=i}^j p_k \Red{\}} & \!i \!\leq\! j.
\end{cases}
$$
This recurrence always recurses on 2 subproblems.
Therefore for every multiedge $(v,U)$ we have $|U|=2$.
The associated multi-DAG has the subproblems/vertices 
$V = \{ (i,j) | 1 \leq i \leq n+1, i-1 \leq j \leq n \}$, 
source $s= (1,n)$ and sinks $\Tcal=\{(i,i-1) | 1\le i\le n+1 \}$.
Also at node $(i,j)$, the set $M^\text{(out)}_{(i,j)}$ consists of $(j-i+1)$ many multiedges.
The $r$th multiedge leaving $(i,j)$ comprised of $2$ edges going from the node $(i,j)$ 
to the nodes $\Green{(i,r-1)}$ and $\Green{(r+1,j)}$.
Figure \ref{fig:obst-dag} illustrates the underlying multi-DAG and the multipath associated with a given BST.

\begin{figure}
\centering
\tiny
\scalebox{1.0}{
\begin{tikzpicture}
\tikzset{vertex/.style = {circle,draw,minimum size=20}}
\tikzset{edge/.style = {->,> = latex, gray!40}}
\tikzset{option1/.style = {->,> = latex, very thick, draw={rgb,255:red,102; green,194; blue,165} }}
\tikzset{option2/.style = {->,> = latex, very thick, draw={rgb,255:red,252; green,141; blue,98} }}
\tikzset{option3/.style = {->,> = latex, very thick, draw={rgb,255:red,141; green,160; blue,203} }}
\tikzset{option4/.style = {->,> = latex, very thick, draw={rgb,255:red,231; green,138; blue,195} }}
\tikzset{option5/.style = {->,> = latex, very thick, draw={rgb,255:red,166; green,216; blue,84} }}

\tikzset{bst/.style = {->,> = latex, blue, very thick}}
\tikzset{rectangle/.style={draw=gray,dashed,fill=green!1,thick,inner sep=5pt}}

\node[rectangle, anchor=south west, minimum width=12cm, minimum height=1cm, rotate around={45:(0,0)}] (A) at (1.5,-.8) {};
\node[text width=1cm] at (6,2) {\large $\Tcal$};
\node[text width=1cm] at (1,8) {\large $s$};

  \foreach \i [evaluate=\i as \imm using int(\i-1)] in {1,...,6}
    \foreach \j in {\imm,...,5} {
    	\node[vertex]  (\i\j) at (1.5*\i,1.5*\j) {(\i,\j)};
    } 

  \foreach \i in {1,...,5} {
    \foreach \j in {\i,...,5} { 
     \foreach \r  [evaluate=\r as \rmm using int(\r-1)]  [evaluate=\r as \rpp using int(\r+1)] in {\i,...,\j} {
		\ifthenelse{\j=\r}{\draw[edge] (\i\j) to (\i\rmm);}{\draw[edge] (\i\j) [bend right] to (\i\rmm);}
		\ifthenelse{\i=\r}{\draw[edge] (\i\j) to (\rpp\j)  ;}{\draw[edge] (\i\j) [bend left] to (\rpp\j);}     
     		}
     	}
     }
     
 	 \node[vertex]   (15) [draw=blue] at (1.5*1,1.5*5) {\textcolor{blue}{(1,5)}};

      \draw[bst] (15) [bend right] to (13) ;
      \draw[bst] (15) [bend left] to (55) ;

 	 \node[vertex]   (15) [draw=blue] at (1.5*5,1.5*5) {\textcolor{blue}{(5,5)}};

	 \draw[bst] (55) to (65);	
	 \draw[bst] (55) to (54);	
	 
 	 \node[vertex]   (15) [draw=blue] at (1.5*6,1.5*5) {\textcolor{blue}{(6,5)}};
 	 \node[vertex]   (15) [draw=blue] at (1.5*5,1.5*4) {\textcolor{blue}{(5,4)}};
 	 
	 \node[vertex]   (13) [draw=blue] at (1.5*1,1.5*3) {\textcolor{blue}{(1,3)}};

      \draw[bst] (13) [bend right] to (10) ;
      \draw[bst] (13)  to (23) ;
      
	 \node[vertex]   (10) [draw=blue] at (1.5*1,1.5*0) {\textcolor{blue}{(1,0)}};
	 
	 \node[vertex]   (23) [draw=blue] at (1.5*2,1.5*3) {\textcolor{blue}{(2,3)}};

      \draw[bst] (23) [bend right] to (21) ;
      \draw[bst] (23)  to (33) ;
      
	 \node[vertex]   (21) [draw=blue] at (1.5*2,1.5*1) {\textcolor{blue}{(2,1)}};
	 
 	 \node[vertex]   (33) [draw=blue] at (1.5*3,1.5*3) {\textcolor{blue}{(3,3)}};

	 \draw[bst] (33) to (43);	
	 \draw[bst] (33) to (32);	
	 
 	 \node[vertex]   (43) [draw=blue] at (1.5*4,1.5*3) {\textcolor{blue}{(4,3)}};
 	 \node[vertex]   (32) [draw=blue] at (1.5*3,1.5*2) {\textcolor{blue}{(3,2)}};

 	 			\tikzset{level distance=10mm, blue ,level/.style={sibling distance=20mm}, draw=very thick}

\tikzstyle{vertex}=[draw=blue,fill=blue!15,circle,minimum
size=20pt,inner sep=0pt]

\node [vertex] at (11, 7) {$k_4$}
    child {
      node [vertex] {$k_1$}
	child[missing]  {}
    child {
      node [vertex] {$k_2$}
	child[missing]  {}
    child {
      node [vertex] {$k_3$}
    }
    }
    }
    child {
      node [vertex] {$k_5$}
    }
;

\end{tikzpicture}
} 

\caption
[Examples of multipaths and multi-DAGs for Binary Search Trees.]
{ 
(Left) An example of a multipath in \Blue{blue} in the underlying multi-DAG.
The nodes in $\Tcal$ represent the subproblems associated  with the ``gaps'' 
e.g.\ $(3,2)$ represents the binary search tree for all values between the keys $2$ and $3$.
(Right) its associated BSTs of $n=5$ keys.
Note that each node, and consequently multiedge, is visited at most once in these multipaths.}
\label{fig:obst-dag}
\end{figure}

Since the above recurrence relation correctly solves the offline optimization problem, 
every multipath in the DAG represents a BST,
and every possible BST can be represented by a multipath of the $2$-DAG.
We have $|M|=O(n^3)$ multiedges which are the components of our new representation. 
The loss of each multiedge leaving $(i,j)$ is $\sum_{k=i}^j p_k $ and is upper bounded by 1.
Most crucially, the original average search cost is linear in the losses of the multiedges and the unit-flow polytope has $O(n^3)$ facets.

\paragraph{Regret Bound.}

As mentioned earlier, the number of binary trees with $n$ nodes is the $n$th Catalan number. 
Therefore  $\Ncal = \frac{(2n)!}{n! (n+1)!} \in (2^n, 4^n)$. 
Also note that each multipath representing a BST consists of exactly $D=n$ multiedges.
Thus using Theorem~\ref{thm:EH-multipaths}, 
EH achieves a regret bound of $\Ocal(n \, \sqrt{L^*})$. 
Moreover, since $|M|=O(n^3)$, using Theorem~\ref{thm:CH-multipaths}, CH achieves a regret bound of
$\Ocal(n^\frac{1}{2} \,(\log n)^{\frac{1}{2}}\, \sqrt{L^*})$. 

\subsection{Matrix-Chain Multiplication}
Given a sequence $A_1, A_2, \ldots, A_n$ of $n$ matrices, 
our goal is to compute the product $A_1 \times A_2 \times \ldots \times A_n$ in the most efficient way.
Using the standard algorithm for multiplying pairs of matrices as a subroutine, 
this product can be found by a specifying the order which the matrices are multiplied together. 
This order is determined by a {\em full parenthesization}:
A product of matrices is fully parenthesized if it is either a single matrix 
or the multiplication of two fully parenthesized matrix
products surrounded by parentheses. For instance, there
are five full parenthesizations of the product $A_1 A_2 A_3 A_4$:
\begin{align*}
&(A_1 (A_2 (A_3 A_4))) \\
&(A_1 ((A_2 A_3) A_4)) \\
&((A_1 A_2) (A_3 A_4)) \\
&(((A_1 A_2) A_3) A_4) \\
&((A_1 (A_2 A_3)) A_4).
\end{align*}

We consider the online version of \textit{matrix-chain multiplication} problem \citep{thomas2001introduction}. 
In each trial, the algorithm predicts with a full
parenthesization of the product 
$A_1 \times A_2 \times \ldots \times A_n$
without knowing the dimensions of these matrices.
Then the adversary reveals the dimensions of each $A_i$ at the end of the trial denoted by $d_{i-1} \times d_i$ for all $i \in \{1..n \}$. 
The loss of the algorithm is defined as the number of scalar multiplications in the matrix-chain product in that trial. 
The goal is to predict with a sequence of full
parenthesizations minimizing regret
which is the difference between the total loss of the algorithm and 
the total loss of the single best full parenthesization chosen in hindsight.
Observe that the number of scalar multiplications in the matrix-chain product
cannot be expressed as a linear loss over the dimensions of the matrices $d_i$'s. 

\paragraph{The Dynamic Programming Representation.}
Finding the best full parenthesization can be solved via dynamic programming \citep{thomas2001introduction}.
Each subproblem is denoted by a pair $(i,j)$ for $1 \leq i \leq j \leq n$, 
indicating the problem of finding a full parenthesization
of the partial matrix product $A_i \ldots A_j$.
The base subproblems are $(i,i)$ for $1 \leq i \leq n$ 
and the complete subproblem is $(1,n)$.
The dynamic programming for matrix chain multiplication uses the following \Red{min-sum} recurrence:
$$\text{OPT}(i,j) = \begin{cases}
0 & i = j \\
\Red{ \min_{i \leq k < j} \{ }\text{OPT}(i , k) \Red{+} \text{OPT}(k+1, j) \Red{+} d_{i-1} \, d_{k} \,d_{j}  \Red{\}} & i < j.
\end{cases}
$$

This recurrence always recurses on 2 subproblems,
thus for all multiedges $m=(v,U)\in M$ we have $|U|=2$.
The associated multi-DAG has the subproblems/vertices 
$V = \{ (i,j) \mid  1 \leq i \leq j \leq n\}$, source
$s=(1,n)$ and sinks $\Tcal = \{ (i,i) \mid 1 \leq i \leq n  \}$.
Also at node $(i,j)$, the set $M^\text{(out)}_{(i,j)}$ consists of $(j-i)$ many multiedges.
The $k$th multiedge leaving $(i,j)$ is comprised of $2$ edges going from the node $(i,j)$ 
to the nodes $(i, k)$ and $(k+1, j)$.
The loss of the $k$th multiedge is $d_{i-1} \, d_{k} \,d_{j}$.
Figure \ref{fig:matrix-dag} illustrates the multi-DAG and multipaths associated with matrix chain multiplications.

\begin{figure}
\centering
\tiny
\scalebox{1.0}{
\begin{tikzpicture}
\tikzset{vertex/.style = {circle,draw,minimum size=20}}
\tikzset{edge/.style = {->,> = latex, gray!40}}

\tikzset{bst/.style = {->,> = latex, blue, very thick}}
\tikzset{rectangle/.style={draw=gray,dashed,fill=green!1,thick,inner sep=5pt}}

\node[rectangle, anchor=south west, minimum width=7.3cm, minimum height=1cm, rotate around={45:(0,0)}] (A) at (1.5,.8) {};
\node[text width=1cm] at (5,3) {\large $\Tcal$};
\node[text width=1cm] at (1,6) {\large $s$};

  \foreach \i [evaluate=\i as \imm using int(\i-1)] in {1,...,4}
    \foreach \j in {\i,...,4} {
    	\node[vertex]  (\i\j) at (1.5*\i,1.5*\j) {(\i,\j)};
    } 

  \foreach \i [evaluate=\i as \ipp using int(\i+1)] in {1,...,3} {
    \foreach \j [evaluate=\j as \jmm using int(\j-1)] in {\ipp,...,4} { 
     \foreach \r  [evaluate=\r as \rmm using int(\r-1)]  [evaluate=\r as \rpp using int(\r+1)] in {\ipp,...,\j} {
		\ifthenelse{\i=\rmm}{\draw[edge] (\i\j) to (\r\j)  ;}{\draw[edge] (\i\j) [bend left] to (\r\j);}     
     		}
     \foreach \r  [evaluate=\r as \rmm using int(\r-1)]  [evaluate=\r as \rpp using int(\r+1)] in {\i,...,\jmm} {
		\ifthenelse{\j=\rpp}{\draw[edge] (\i\j) to (\i\r);}{\draw[edge] (\i\j) [bend right] to (\i\r);}
     		}
     	}
     }
     
 	 \node[vertex]   (14) [draw=blue] at (1.5*1,1.5*4) {\textcolor{blue}{(1,4)}};
 	 \node[vertex]   (24) [draw=blue] at (1.5*2,1.5*4) {\textcolor{blue}{(2,4)}};
 	 \node[vertex]   (23) [draw=blue] at (1.5*2,1.5*3) {\textcolor{blue}{(2,3)}};
 	 \node[vertex]   (11) [draw=blue] at (1.5*1,1.5*1) {\textcolor{blue}{(1,1)}};
 	 \node[vertex]   (22) [draw=blue] at (1.5*2,1.5*2) {\textcolor{blue}{(2,2)}};
 	 \node[vertex]   (33) [draw=blue] at (1.5*3,1.5*3) {\textcolor{blue}{(3,3)}};
 	 \node[vertex]   (44) [draw=blue] at (1.5*4,1.5*4) {\textcolor{blue}{(4,4)}};

	 \draw[bst] (14) [bend right] to (11);	
	 \draw[bst] (14) to (24);	

	 \draw[bst] (24) to (23);	
	 \draw[bst] (24) [bend left] to (44);	
	 
	 \draw[bst] (23) to (22);	
	 \draw[bst] (23) to (33);	

\end{tikzpicture}
} 
\caption[Examples of multipaths and multi-DAGs for Matrix-Chain Multiplications.]
{Given a chain of $n=4$ matrices, the
multipath associated with the full parenthesization
\textcolor{blue}{$(A_1((A_2 A_3)A_4))$} 
is depicted in \textcolor{blue}{blue}.}
\label{fig:matrix-dag}
\end{figure}

Since the above recurrence relation correctly solves the offline optimization problem, 
every multipath in the multi-DAG represents a full parenthesization,
and every possible full parenthesization can be represented by a
 multipath of the multi-DAG.
We have $|M|=O(n^3)$ multiedges which are the components of our new representation. 
Assuming that all dimensions $d_i$ are bounded as $d_i < d_\text{max}$ for some $d_\text{max}$, 
the loss associated with each multiedge is upper-bounded by $(d_\text{max})^3$.
Most crucially,
the original number of scalar multiplications in the matrix-chain product is linear in the losses of
the multiedges and the unit-flow polytope has $O(n^3)$ facets.

\paragraph{Regret Bounds.}
It is well-known that the number of full parenthesizations
of a sequence of $n$ matrices is the $n$th Catalan number
\citep{thomas2001introduction}. 
Therefore  $\Ncal = \frac{(2n)!}{n! (n+1)!} \in (2^n, 4^n)$. 
Also note that each multipath representing a full parenthesization consists of exactly $D=n-1$ multiedges.
Thus, incorporating $(d_\text{max})^3$ as the loss range for each component and using Theorem~\ref{thm:EH-multipaths}, 
EH achieves a regret bound of $\Ocal(n \, (d_\text{max})^\frac{3}{2} \sqrt{L^*})$. 
Moreover, since $|M|=O(n^3)$, using Theorem~\ref{thm:CH-multipaths} and considering $(d_\text{max})^3$ as the loss range for each component , 
CH achieves a regret bound of $\Ocal(n^\frac{1}{2} \,(\log n)^{\frac{1}{2}}\, (d_\text{max})^\frac{3}{2} \sqrt{L^*})$.

\subsection{Knapsack}
Consider the online version of the {\em knapsack} problem \citep{kleinberg2006algorithm}:
We are given a set of $n$ items along with the {\em capacity} of the knapsack $C \in \NN$. 
For each item $i \in \{1..n\}$, a {\em heaviness} $h_i \in \NN$ is associated.
In each trial, the algorithm predicts with a \textit{packing} which is a subset of items whose total heaviness is at most the capacity of the knapsack.
After the prediction of the algorithm, 
the adversary reveals the profit of each item $p_i \in [0,1]$.
The gain is defined as the sum of the profits of the items picked in the packing predicted by the algorithm in that trial. 
The goal is to predict with a sequence of packings minimizing regret
which is the difference between the total gain of the algorithm and 
the total gain of the single best packing chosen in hindsight.

Note that this online learning problem only deals with exponentially many solutions when 
there are exponentially many feasible packings. 
If the number of packings is polynomial, then it is practical to simply run the Hedge algorithm 
with one weight per packing.
Here we consider a setting of the problem where maintaining one weight per packing is impractical.
We assume $C$ and $h_i$'s are in such way that the number of feasible packings is exponential in $n$.

\paragraph{The Dynamic Programming Representation.}
Finding the optimal packing can be solved via dynamic programming \citep{kleinberg2006algorithm}.
Each subproblem is denoted by a pair $(i,c)$ for $0 \leq i \leq n$ and $0 \leq c \leq C$, 
indicating the knapsack problem given items $1,\ldots,i$ and capacity $c$. 
The base subproblems are $(0,c)$ for $0 \leq c \leq C$
and the complete subproblem is $(n, C)$.
The dynamic programming for the knapsack problem uses the following \Red{max-sum} recurrence:
$$\text{OPT}(i,c) = \begin{cases}
0 & i = 0 \\
\text{OPT}(i -1, c) & c<h_i \\
\Red{ \max \{ }\text{OPT}(i - 1, c),\:\: p_i \Red{+} \text{OPT}(i -1, c-h_i) \Red{\}} & \text{else}.
\end{cases}
$$

This recurrence always recurses on 1 subproblem.
Thus the multipaths are regular paths 
and the problem is essentially the online longest-path problem with several sink nodes.
The associated DAG has the subproblems/vertices 
$V = \{ (i,c) \mid 0 \leq i \leq n, \quad 0 \leq c \leq C \}$, source $s=(n,C)$ and sinks 
$\Tcal = \{ (0,c) \mid  0 \leq c \leq C \}$.
Also at node $(i,c)$, the set $M^\text{(out)}_{(i,c)}$ consists of two edges going from 
the node $(i,c)$ to the nodes $(i-1, c)$ and $(i-1, c-h_i)$.
Figure \ref{fig:knapsack-dag} illustrates an example of the DAG 
and a sample path associated with a packing.

\begin{figure}
\centering
\tiny
\begin{tikzpicture}
\tikzset{vertex/.style = {circle,draw,minimum size=20}}
\tikzset{edge/.style = {->,> = latex}}
\tikzset{packing/.style = {->,> = latex, very thick, blue}}
\tikzset{rectangle/.style={draw=gray,dashed,fill=green!1,thick,inner sep=5pt}}

\node[rectangle, anchor=south west, minimum width=11.5cm, minimum height=1cm] (A) at (-.5,-.5) {};
\node[text width=1cm] at (-.5,.5) {\large $\Tcal$};
\node[text width=1cm] at (11.3,5.1) {\large $s$};

  \foreach \x in {0,...,7}
    \foreach \y in {0,...,3} 
       {
       \node[vertex]  (\x\y) at (1.5*\x,1.5*\y) {(\y,\x)};} 


  \foreach \x in {0,...,7}
    \foreach \y [count=\yi] in {0,1,2}  
      \draw[edge] (\x\yi) to["$0$"] (\x\y)  ; 

\foreach \l [count=\li from 0] in {2,3,4}{
\pgfmathtruncatemacro{\U}{7- \l}
  \foreach \x in {0,...,\U}{
    \foreach \y [count=\yi from \li+1] in {\li}  {
      {\pgfmathtruncatemacro{\newx}{\x + \l}
      \draw[edge]  (\newx\yi) to["$p_\yi$" {pos=0.2}] (\x\y)  ; 
      }
    }
  }
}



	 \node[vertex]  (73) [draw=blue] at (1.5*7,1.5*3) {\textcolor{blue}{(3,7)}}; 
	\draw[packing] (73) to (32);
	
	 \node[vertex]  (32) [draw=blue] at (1.5*3,1.5*2) {\textcolor{blue}{(2,3)}}; 
	\draw[packing] (32) to (31);

	 \node[vertex]  (31) [draw=blue] at (1.5*3,1.5*1) {\textcolor{blue}{(1,3)}}; 
	 \node[vertex]  (10) [draw=blue] at (1.5*1,1.5*0) {\textcolor{blue}{(0,1)}}; 

	\draw[packing] (31) to (10);

	
\end{tikzpicture}
\caption[Examples of paths and DAGs for the Knapsack problem.]
{ An example with $C = 7$ and $(h_1, h_2, h_3) = (2, 3, 4)$. The packing of picking the first and third item is highlighted.}
\label{fig:knapsack-dag}
\end{figure}

Since the above recurrence relation correctly solves the offline optimization problem, 
every path in the DAG represents a packing, 
and every possible packing can be represented by a path of the DAG.
We have $|M|=|E|=O(n\, C)$ edges which are the components of our new representation. 
The gains of the edges going from the node $(i,c)$ to the nodes $(i-1, c)$ and $(i-1, c-h_i)$ 
are $0$ and $p_i$, respectively.
Note that the gain associated with each edge is upper-bounded by $1$. 
Most crucially,
the sum of the profits of the picked items in the packing is linear in the gains of
the edges and the unit-flow polytope has $O(n \, C)$ facets.

\paragraph{Regret Bounds.}
We turn the problem into an equivalent shortest-path problem 
by defining a loss for each edge $e \in E$ as $\ell_e = 1 - g_e$ in which $g_e$ is the gain of $e$. 
Call this new DAG $\bar{\Gcal}$. 
Let $L_{\bar{\Gcal}}(\pi)$ be the loss of path $\pi$ in
$\bar{\Gcal}$ and $G_{{\Gcal}}(\pi)$ be the gain of path $\pi$ in
$\Gcal$.
Since all paths contain exactly $D=n$ edges, 
the loss and gain are related as follows:
$L_{\bar{\Gcal}}(\pivec) = n - G_{{\Gcal}}(\pivec)$.
According to our initial assumption $\log \Ncal = \Ocal( n )$.
Thus using Theorem \ref{thm:EH-multipaths} we obtain:
\begin{align*}
G^* - \EE[G_\text{EH}] 
&= (nT - L^*) - (nT - \EE[L_\text{EH}]) \\
&= \EE[L_\text{EH}] - L^* 
= \Ocal( n  \, \sqrt{L^*} ).
\end{align*}

Notice that the number of multiedges/edges is $|M|=|E|=O(n\, C)$ and each path consists of $D=n$ edges. Therefore using Theorem \ref{thm:CH-multipaths} we obtain:
\begin{align*}
G^* - \EE[G_\text{CH}] 
&= (nT - L^*) - (nT - \EE[L_\text{CH}]) \\
&=\EE[L_\text{CH}] - L^* 
= \Ocal( n^\frac{1}{2} \, (\log nC)^\frac{1}{2} \, \sqrt{L^*} ).
\end{align*}

\subsection{$k$-Sets}

Consider the online learning of the {\em $k$-sets} \citep{warmuth2008randomized}:
We want to learn subsets of size $k$ of the set $\{1..n\}$.
In each trial, the algorithm predicts with a $k$-set.
Then, the adversary reveals the loss of each element $\ell_i$ for $i \in \{1..n\}$.
The loss is defined as the sum of the losses of the elements in the $k$-set predicted by the algorithm in that trial. 
The goal is to predict with a sequence of $k$-sets minimizing regret
which is the difference between the total loss of the algorithm and 
the total loss of the single best $k$-set chosen in hindsight.

\paragraph{The Dynamic Programming Representations.}

Finding the optimal $k$-set can be solved via dynamic programming.
Each subproblem is denoted by a pair $(i,j)$ for $0 \leq j \leq k$ and $j \leq i \leq j + n - k$,
indicating the $j$-set problem over the set $\{1,\ldots,i\}$.
The base subproblem is $(0,0)$ and the complete subproblem is $(n,k)$.
The dynamic programming for the $k$-set problem uses the following \Red{min-sum} recurrence:

$$\text{OPT}(i,j) = 
\begin{cases}
0 & i=j=0 \\
\text{OPT}(i-1,0) & j=0 \\
\text{OPT}(i-1,i-1) \Red{+} \ell_i & j=i \\
\Red{\min \{ } \text{OPT}(i-1,j) \Red{,}  \text{OPT}(i-1,j-1) \Red{+} \ell_i \Red{\}} & \text{otherwise}.
\end{cases}
$$

This recurrence always recurses on 1 subproblem.
Thus the multipaths are regular paths 
and the problem is essentially the online shortest-path problem from a source to a sink.
The associated DAG has the subproblems/vertices 
$V = \{ (i,j) \mid 0 \leq j \leq k, \quad j \leq i \leq j + n - k \}$, 
source $s=(n,k)$ and sink $\Tcal = \{ (0,0)  \}$.
Also at node $(i,j)$, the set $M^\text{(out)}_{(i,j)}$ consists of two edges going from 
the node $(i,j)$ to the nodes $(i-1, j)$ and $(i-1, j-1)$.
Figure \ref{fig:kset-dag} illustrates an example of the DAG 
and a sample path associated with a $k$-set.

\begin{figure}
\centering
\tiny
\begin{tikzpicture}
\tikzset{vertex/.style = {circle,draw,minimum size=20}}
\tikzset{edge/.style = {->,> = latex}}
\tikzset{packing/.style = {->,> = latex, very thick, blue}}
\tikzset{rectangle/.style={draw=gray,dashed,fill=green!1,thick,inner sep=5pt}}

\node[rectangle, anchor=south west, minimum width=1cm, minimum height=1cm] (A) at (-.5,-.5) {};
\node[text width=1cm] at (-.5,.5) {\large $\Tcal$};
\node[text width=1cm] at (7,5.1) {\large $s$};

  \foreach \x in {0,...,4}
    \foreach \y in {0,...,3} 
       {\pgfmathtruncatemacro{\newx}{\x +  \y }
       \node[vertex]  (\x\y) at (1.5*\x,1.5*\y) {(\newx,\y)};} 


  \foreach \x [count=\xi] in {0,...,4}
    \foreach \y [count=\yi] in {0,1,2}  
    	{\pgfmathtruncatemacro{\newx}{\x +  \y + 1}
      \draw[edge] (\x\yi) to["$\ell_\newx$"] (\x\y)  ; }

  \foreach \x [count=\xi] in {0,...,3}
    \foreach \y [count=\yi] in {0,1,2,3}  
    	{
      \draw[edge] (\xi\y) to["$0$"] (\x\y)  ;  }

	 \node[vertex]  (43) [draw=blue] at (1.5*4,1.5*3) {\textcolor{blue}{(7,3)}}; 
	\draw[packing] (43) to (33);
	
	 \node[vertex]  (33) [draw=blue] at (1.5*3,1.5*3) {\textcolor{blue}{(6,2)}}; 
	\draw[packing] (33) to (23);

	 \node[vertex]  (23) [draw=blue] at (1.5*2,1.5*3) {\textcolor{blue}{(5,2)}}; 
	\draw[packing] (23) to (22);
	
	 \node[vertex]  (22) [draw=blue] at (1.5*2,1.5*2) {\textcolor{blue}{(4,2)}}; 
	\draw[packing] (22) to (21);
	
	 \node[vertex]  (21) [draw=blue] at (1.5*2,1.5*1) {\textcolor{blue}{(3,1)}}; 
	\draw[packing] (21) to (11);
	
	 \node[vertex]  (11) [draw=blue] at (1.5*1,1.5*1) {\textcolor{blue}{(2,1)}}; 
	\draw[packing] (11) to (01);
	
	 \node[vertex]  (01) [draw=blue] at (1.5*0,1.5*1) {\textcolor{blue}{(1,0)}}; 
	\draw[packing] (01) to (00);

	 \node[vertex]  (00) [draw=blue] at (1.5*0,1.5*0) {\textcolor{blue}{(0,0)}};

	
\end{tikzpicture}
\caption[Examples of paths and DAGs for the $k$-set problem.]
{ 
An example of $k$-set with $n=7$ and $k=3$. 
The $3$-set of $(1,0,0,1,1,0,0)$ is highlighted.
}
\label{fig:kset-dag}
\end{figure}

Since the above recurrence relation correctly solves the offline $k$-set problem, 
every path in the DAG represents a $k$-set, 
and every possible $k$-set can be represented by a path of the DAG.
We have $|M|=|E|=2k(n-k)+n$ edges which are the components of our new representation. 
The losses of the edges going from the node $(i,j)$ to the nodes $(i-1, j)$ and $(i-1, j-1)$ 
are $0$ and $\ell_i$, respectively.
Note that the loss associated with each edge is upper-bounded by $1$. 
Most crucially,
the sum of the losses of the predicted $k$-set is linear in the losses of
the edges and the unit-flow polytope has $O(k (n-k))$ facets.

\paragraph{Regret Bounds.}
The number of $k$-sets is $\Ncal = {n \choose k}$.
Also note that each path representing a $k$-set consists of exactly $D=n$ edges and its loss is bounded by $k$.
Thus, using Theorem~\ref{thm:EH-multipaths}, EH achieves a regret bound of $\Ocal(k \, (\log n)^\frac{1}{2} \sqrt{L^*})$. 
Moreover, since $|E|=O(k(n-k))$, using Theorem~\ref{thm:CH-multipaths},
CH achieves a regret bound of $\Ocal(n^\frac{1}{2} \,(\log k(n-k))^{\frac{1}{2}} \sqrt{L^*})$.

\paragraph{Remark.}
The convex hull of the $k$-sets in its original space, 
known as \emph{capped probability simplex}, is well-behaved.
This polytope has $n+1$ facets and the exact relative entropy projection to this polytope can be found efficiently \citep{warmuth2008randomized}.
Thus applying CH in the original space will result in more efficient algorithm with better bounds of $\Ocal(k^\frac{1}{2} \, (\log n)^\frac{1}{2} \sqrt{L^*})$. 
Nevertheless, an efficient implementation of the EH algorithm can be obtained via our
online dynamic programming framework.
%
Interestingly, in the special case of the $k$-set, the regret bounds of EH is also $\Ocal(k^\frac{1}{2} \, (\log n)^\frac{1}{2} \sqrt{L^*})$ \citep{kivinen2010kset}.

\subsection{Rod Cutting}
\begin{figure}
\centering
\includegraphics[width=.7\textwidth]{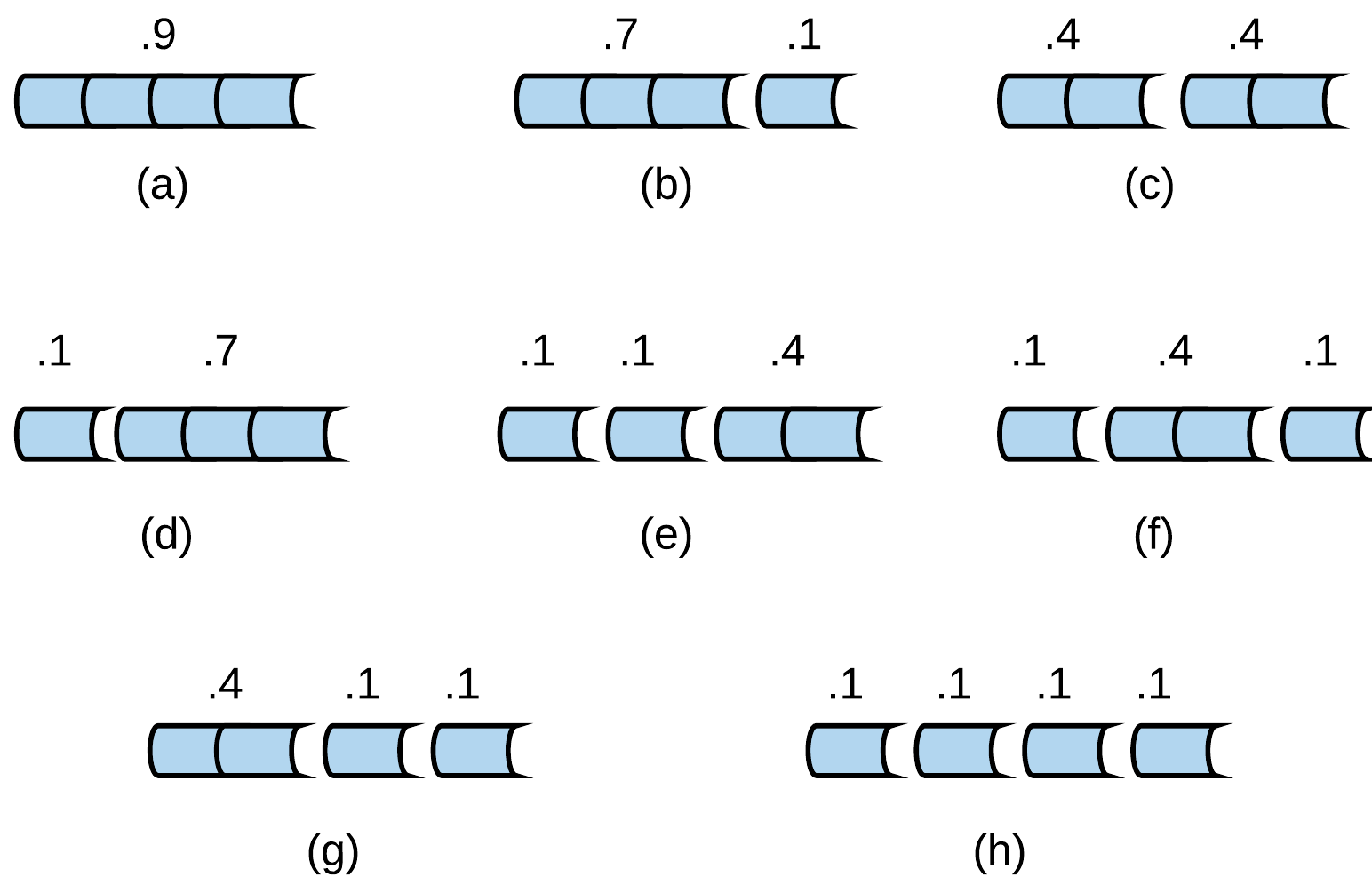}
\caption[An example of all cutting for a given rod in the rod cutting problem.]
{All cuttings of a rod of length $n=4$ and their profits given  $(p_1, p_2, p_3, p_4) = (.1, .4, .7, .9)$. }
\label{fig:rod-ex}
\end{figure}

Consider the online version of {\em rod cutting} problem \citep{thomas2001introduction}:
A rod of length $n \in \NN$ is given.  
In each trial, the algorithm predicts with a \textit{cutting}, that is, it cuts up the rod into smaller pieces of integer length. 
Then the adversary reveals a {\em profit} $p_i \in [0,1]$ for each piece of length $i \in \{1..n \}$ that can be possibly generated out of a cutting.
The gain of the algorithm is defined as the sum of the profits of all the pieces generated by the predicted cutting in that trial. 
The goal is to predict with a sequence of cuttings minimizing regret
which is the difference between the total gain of the algorithm and 
the total gain of the single best cutting chosen in hindsight. 
See Figure \ref{fig:rod-ex} as an example. 

\paragraph{The Dynamic Programming Representation.}
Finding the optimal cutting can be solved via dynamic programming \citep{thomas2001introduction}.
Each subproblem is simply denoted by $i$ for $0 \leq i \leq n$, 
indicating the rod cutting problem given a rod of length $i$.
The base subproblem is $i=0$, and the complete subproblem is $i=n$.
The dynamic programming for the rod cutting problem uses the following \Red{max-sum} recurrence:
$$\text{OPT}(i) = 
\begin{cases}
0 & i=0 \\
\Red{\max_{0 \leq j \leq i} \{ } OPT(j) \Red{+} p_{i-j} \Red{\}}& i>0.
\end{cases}
$$

This recurrence always recurses on 1 subproblem.
Thus the multipaths are regular paths 
and the problem is essentially the online longest-path problem from the source to the sink.
The associated DAG has the subproblems/vertices 
$V = \{ 0, 1, \ldots, n \}$, source $s=n$ and sink $\Tcal = \{ 0 \}$.
Also at node $i$, the set $M^\text{(out)}_i$ consists of $i$ edges going from 
the node $i$ to the nodes $0, 1, \ldots, i-1$.
Figure \ref{fig:rod-dag} illustrates the DAG and paths associated with the cuttings.

\begin{figure}
\centering
\begin{tikzpicture}%
  [>=stealth,
   shorten >=1pt,
   node distance=2cm,
   on grid,
   auto,
   every state/.style={draw=black!60, fill=black!5, very thick}
  ]
  
\tikzset{cutting/.style = {->,> = latex, very thick, blue}}
    \tikzset{rectangle/.style={draw=gray,dashed,fill=green!1,thick,inner sep=5pt}}

\node[rectangle, anchor=south west, minimum width=1.3cm, minimum height=1.2cm] (A) at (-.6,-.6) {};
\node[text width=1cm] at (0,1.2) {\large $\Tcal$};
\node[text width=1cm] at (9,.5) {\large $s$};

\node[state] (0)                  {0};
\node[state] (1) [right=of 0] {1};
\node[state] (2) [right=of 1] {2};
\node[state] (3) [right=of 2] {3};
\node[state] (4) [right=of 3] {4};

\path[->]
   (4) edge     node                      {$p_1$} (3)
         edge[bend right] node [above] {$p_2$} (2)
         edge[bend right] node [above] {$p_3$} (1)
         edge[bend right] node [above] {$p_4$} (0)
   (3) edge     node                      {$p_1$} (2)
         edge[bend right] node [above] {$p_2$} (1)
         edge[bend right] node [above] {$p_3$} (0)
   (2) edge     node                      {$p_1$} (1)
         edge[bend right] node [above] {$p_2$} (0)
   (1) edge     node                      {$p_1$} (0)
   ;
   
\draw[cutting] (4) [bend right] to (2);
\draw[cutting] (2) [bend right] to (0);

\end{tikzpicture}
\caption[Examples of paths and DAGs for the rod cutting problem.]
{An example of rod cutting problem with $n=4$. The cutting with two smaller pieces of size $2$ is highlighted.}
\label{fig:rod-dag}
\end{figure}

Since the above recurrence relation correctly solves the offline optimization problem, 
every path in the DAG represents a cutting, 
and every possible cutting can be represented by a path of the DAG.
We have $|M|=|E|=O(n^2)$ multiedges/edges which are the components of our new representation. 
The gains of the edges going from the node $i$ to the node $j$ (where $j<i$) is $p_{i-j}$.
Note that the gain associated with each edge is upper-bounded by $1$. 
Most crucially,
the sum of the profits of all the pieces generated by the cutting is linear in the gains of
the edges and the unit-flow polytope has $O(n)$ facets.

\paragraph{Regret Bounds.}
Similar to the knapsack problem, we turn this problem into
a shortest-path problem:
We first modify the graph so that all paths have equal length of $n$ 
(which is the length of the longest path) 
and the gain of each path remains fixed.
We apply a method introduced in \cite{gyorgy2007line}, 
which adds $\Ocal( n^2 )$ vertices and edges (with gain
zero) to make all paths have the same length of $D=n$. 
Then we define a loss for each edge $e$ as $\ell_e = 1 - g_e$ in which $g_e$ is the gain of $e$.
Call this new DAG $\bar{\Gcal}$. Similar to the knapsack problem, we have $L_{\bar{\Gcal}}(\pivec) = n - G_{{\Gcal}}(\pivec)$ for all paths $\pivec$.
Note that in both $\Gcal$ and $\bar{\Gcal}$, there are $\Ncal = 2^{n-1}$ paths.
Thus using Theorem \ref{thm:EH-multipaths} we obtain%
\footnote{We are over-counting the number of cuttings. The number of possible cutting is called {\em partition function}
which is approximately $e^{\pi \sqrt{2n/3}} / 4 n \sqrt{3}$ \citep{thomas2001introduction}. 
Thus if we run the Hedge algorithm inefficiently with one weight per cutting, 
we will get a better regret bound by
a factor of $\sqrt[4]{n}$.
}
\begin{align*}
G^* - \EE[G_\text{EH}] 
&= (nT - L^*) - (nT - \EE[L_\text{EH}]) \\
&= \EE[L_\text{EH}] - L^* 
= \Ocal( n \, \sqrt{L^*} ).
\end{align*}

Notice that the number of multiedges/edges in $\bar{\Gcal}$ is $|M|=|E|=O(n^2)$  and each path consists of $D=n$ edges. 
Therefore using Theorem \ref{thm:CH-multipaths} we obtain:
\begin{align*}
G^* - \EE[G_\text{CH}] 
&= (nT - L^*) - (nT - \EE[L_\text{CH}]) \\
&=\EE[L_\text{CH}] - L^* 
= \Ocal( n^\frac{1}{2} \, (\log n)^\frac{1}{2} \, \sqrt{L^*} ).
\end{align*}

\subsection{Weighted Interval Scheduling}
\begin{figure}
\centering
\includegraphics[width=.7\textwidth]{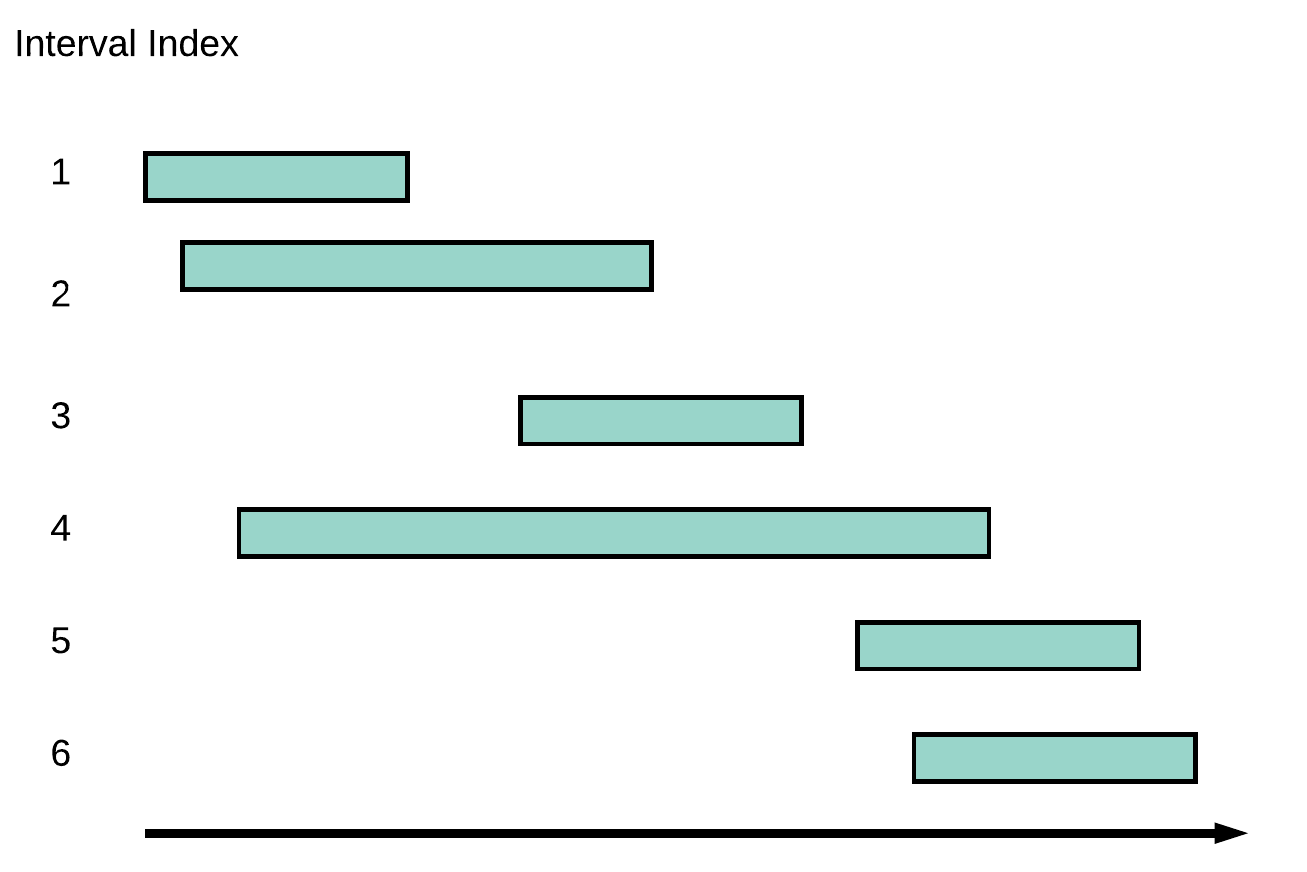}
\caption{An example of weighted interval scheduling with $n=6$}
\label{fig:interval-ex}
\end{figure}

Consider the online version of \textit{weighted interval scheduling} problem \citep{kleinberg2006algorithm}:
We are given a set of $n$ intervals $I_1, \ldots, I_n$ on the real line. 
In each trial, the algorithm predicts with a {\em scheduling} which is a subset of non-overlapping intervals. 
Then, for each interval $I_j$, the adversary reveals $p_j \in [0,1]$ which is the {\em profit} of including $I_j$ in the scheduling.
The gain of the algorithm is defined as the total profit over chosen intervals in the scheduling in that trial. 
The goal is to predict with a sequence of schedulings minimizing regret
which is the difference between the total gain of the algorithm and 
the total gain of the single best scheduling chosen in hindsight. 
See Figure \ref{fig:interval-ex} as an example.
Note that this problem is only interesting when 
the number of solutions (i.e. schedulings) are exponential in $n$.

\paragraph{The Dynamic Programming Representation.}
Finding the optimal scheduling can be solved via dynamic programming \citep{kleinberg2006algorithm}.
Each subproblem is simply denoted by $i$ for $0 \leq i \leq n$, 
indicating the weighted scheduling problem for the intervals $I_1, \ldots, I_i$.
The base subproblem is $i=0$, and the complete subproblem is $i=n$.
The dynamic programming for the weighted interval scheduling problem uses the following \Red{max-sum} recurrence:
$$\text{OPT}(i) = 
\begin{cases}
0 & i=0 \\
\Red{\max \{ }\text{OPT}(i-1), \text{OPT}(\text{pred}(i)) \Red{+} p_{i} \Red{\}} & i>0.
\end{cases}
$$
where
$$
\text{pred}(i):= \begin{cases}
0 & i=1 \\
\max_{ \{ j<i, \, I_i \cap I_j = \emptyset \}}  \; j & i > 1.
\end{cases}$$

This recurrence always recurses on 1 subproblem.
Thus the multipaths are regular paths 
and the problem is essentially the online longest-path problem from the source to the sink.
The associated DAG has the subproblems/vertices 
$V = \{ 0, 1, \ldots, n \}$, source $s=n$ and sink $\Tcal = \{ 0 \}$.
Also at node $i$, the set $M^\text{(out)}_i$ consists of $2$ edges going from 
the node $i$ to the nodes $i-1$ and pred$(i)$.
Figure \ref{fig:interval-dag} illustrates the DAG and paths
associated with the scheduling for 
the example given in Figure \ref{fig:interval-ex}.

\begin{figure}
\centering
\begin{tikzpicture}%
  [>=stealth,
   shorten >=1pt,
   node distance=1.5cm,
   on grid,
   auto,
   every state/.style={draw=black!60, fill=black!5, very thick}
  ]
  
\tikzset{scheduling/.style = {->,> = latex, very thick, blue}}
    \tikzset{rectangle/.style={draw=gray,dashed,fill=green!1,thick,inner sep=5pt}}

\node[rectangle, anchor=south west, minimum width=1.3cm, minimum height=1.2cm] (A) at (-.6,-.6) {};
\node[text width=1cm] at (0,1.2) {\large $\Tcal$};
\node[text width=1cm] at (10,.5) {\large $s$};

\node[state] (0)                  {0};
\node[state] (1) [right=of 0] {1};
\node[state] (2) [right=of 1] {2};
\node[state] (3) [right=of 2] {3};
\node[state] (4) [right=of 3] {4};
\node[state] (5) [right=of 4] {5};
\node[state] (6) [right=of 5] {6};

\path[->]
   (6) edge     node                      {$0$} (5)
         edge[bend right] node [above] {$p_6$} (3)
   (5) edge     node                      {$0$} (4)
         edge[bend right] node [above] {$p_5$} (3)
   (4) edge     node                      {$0$} (3)
         edge[bend right] node [above] {$p_4$} (0)
   (3) edge     node                      {$0$} (2)
         edge[bend right] node [above] {$p_3$} (1)
   (2) edge     node                      {$0$} (1)
         edge[bend right] node [above] {$p_2$} (0)
   (1) edge     node                      {$0$} (0)
         edge[bend right] node [above] {$p_1$} (0)
   ;
   
\draw[scheduling] (6)  to (5);
\draw[scheduling] (5) [bend right] to (3);
\draw[scheduling] (3) [bend right] to (1);
\draw[scheduling] (1)  [bend right] to (0);

\end{tikzpicture}
\caption[Examples of paths and DAG for the weight interval scheduling problem.]
{ The underlying DAG associated with the example illustrated in Figure \ref{fig:interval-ex}. The scheduling with $I_1$, $I_3$, and $I_5$ is highlighted.}
\label{fig:interval-dag}
\end{figure}

Since the above recurrence relation correctly solves the offline optimization problem, 
every path in the DAG represents a scheduling, 
and every possible scheduling can be represented by a path of the DAG.
We have $|M|=|E|=O(n)$ multiedges/edges which are the components of our new representation. 
The gains of the edges going from the node $i$ to the nodes $i-1$ and pred$(i)$
are $0$ and $p_i$, respectively.
Note that the gain associated with each edge is upper-bounded by $1$. 
Most crucially,
the total profit over chosen intervals in the scheduling is linear in the gains of
the edges and the unit-flow polytope has $O(n)$ facets.


\paragraph{Regret Bounds.}
Similar to rod cutting, this is also the online longest-path problem with one sink node.
Like the rod cutting problem, we modify the graph by adding
$\Ocal(n^2)$ vertices and edges (with gain zero) to make
all paths have the same length of $D=n$ and change the gains into losses.
Call this new DAG $\bar{\Gcal}$. Again we have $L_{\bar{\Gcal}}(\pivec) = n - G_{{\Gcal}}(\pivec)$ for all paths $\pivec$.
According to our initial assumption $\log \Ncal = \Ocal( n )$.
Thus using Theorem \ref{thm:EH-multipaths} we obtain:
\begin{align*}
G^* - \EE[G_\text{EH}] 
&= (nT - L^*) - (nT - \EE[L_\text{EH}]) \\
&= \EE[L_\text{EH}] - L^* 
= \Ocal( n \, \sqrt{L^*} ).
\end{align*}

Notice that the number of multiedges/edges  in $\bar{\Gcal}$ is $|M|=|E|=O(n^2)$  and each path consists of $D=n$ edges. 
Therefore using Theorem \ref{thm:CH-multipaths} we obtain:
\begin{align*}
G^* - \EE[G_\text{CH}] 
&= (nT - L^*) - (nT - \EE[L_\text{CH}]) \\
&=\EE[L_\text{CH}] - L^* 
= \Ocal( n^\frac{1}{2} \, (\log n)^\frac{1}{2} \, \sqrt{L^*} ).
\end{align*}


\section{Conclusions and Future Work}
\label{sec:odp-conclusions}

\begin{table}
\centering
\begin{tabular}{| c | c | c | c |}
\hline
Problem & FPL & EH &  CH \\
\hline
Optimal Binary 
& $\Ocal( n\, (\log n)^{\frac{1}{2}} \, \sqrt{L^*} )$ 
& $\Ocal( n \, \sqrt{L^*} )$ 
& $\Ocal( n^{\frac{1}{2}} \, (\log n)^{\frac{1}{2}} \, \sqrt{L^*} )$ \\

Search Trees
& ~
& ~
& \Purple{\textbf{*Best*}} \\

\hline

Matrix-Chain 
& ---
& $\Ocal( n \, (d_\text{max})^{\frac{3}{2}} \, \sqrt{L^*} )$ 
& $\Ocal( n^\frac{1}{2} \, (\log n)^{\frac{1}{2}} \, (d_\text{max})^\frac{3}{2} \, \sqrt{L^*} )$ \\

Multiplications \footnotemark
& ~
& ~
& \Purple{\textbf{*Best*}} \\

\hline

Knapsack
& $\Ocal( n\, (\log n)^{\frac{1}{2}} \, \sqrt{L^*} )$ 
& $\Ocal( n \, \sqrt{L^*} )$ 
& $\Ocal( n^{\frac{1}{2}} \, (\log nC)^{\frac{1}{2}} \, \sqrt{L^*} )$ \\

& & & \Purple{\textbf{*Best*}}\\

\hline

$k$-sets
& $\Ocal( k^\frac{1}{2} \, n^\frac{1}{2} \, (\log n)^\frac{1}{2} \sqrt{L^*} )$ 
& $\Ocal( k^\frac{1}{2} \, (\log n)^\frac{1}{2} \sqrt{L^*} )$ 
& $\Ocal(n^\frac{1}{2} \,(\log k(n-k))^{\frac{1}{2}} \sqrt{L^*})$ \\

& & \Purple{\textbf{*Best*}} & \\

\hline

Rod Cutting 
& $\Ocal( n\, (\log n)^{\frac{1}{2}} \, \sqrt{L^*} )$ 
& $\Ocal( n \, \sqrt{L^*} )$ 
& $\Ocal( n^{\frac{1}{2}} \, (\log n)^{\frac{1}{2}} \, \sqrt{L^*} )$ \\

& & & \Purple{\textbf{*Best*}}\\

\hline

Weighted Interval  
& $\Ocal( n\, (\log n)^{\frac{1}{2}} \, \sqrt{L^*} )$ 
& $\Ocal( n \, \sqrt{L^*} )$ 
& $\Ocal( n^{\frac{1}{2}} \, (\log n)^{\frac{1}{2}} \, \sqrt{L^*} )$ \\

Scheduling
& ~
& ~
& \Purple{\textbf{*Best*}} \\

\hline
\end{tabular}
\caption
[Performance of various algorithms over different problems in the full information setting.]
{Performance of various algorithms over different problems in the full information setting. $C$ is the capacity in the Knapsack problem, and $d_\text{max}$ is the upper-bound on the dimension in matrix-chain multiplication problem.}
\label{table:regret-comparison}
\end{table}
\footnotetext{The loss of a fully parenthesized matrix-chain multiplication 
is the number of scalar multiplications in the execution of
all matrix products.
This number cannot be expressed as a linear loss over the dimensions 
of the matrices. 
We are thus unaware of a way to apply FPL to this problem 
using the dimensions of the matrices as the components. }

We developed a general framework for combinatorial online learning problmes
whose offline optimization problems can be efficiently solved via 
``min-sum'' dynamic programming algorithms. 
Table~\ref{table:regret-comparison} gives the performance
of EH and CH in our dynamic programming framework
and compares it with the Follow the Perturbed Leader (FPL)
algorithm \citep{kalai2005efficient}. FPL additively perturbs the losses and then uses
dynamic programming to find the solution of minimum loss.
FPL is always worse than EH and CH. 
CH is better than both FPL and EH in all cases except $k$-set.
In the case of $k$-sets, CH can be better implemented 
in the original space by using the capped probability simplex 
as the polytope \citep{warmuth2008randomized, koolen2010hedging}
rather than the dynamic programming representation 
and the unit-flow polytope.

We conclude with a few remarks:
\begin{itemize} 
\item For EH, projections are simply a renormalization of
the weight vector.
In contrast, iterative Bregman projections are often needed
for projecting back into the polytope used by CH \citep{koolen2010hedging, helmbold2009learning}.
These methods are known to converge to the exact projection 
\citep{bregman1967relaxation, bauschke1997legendre}
and are reported to be very efficient empirically \citep{koolen2010hedging}.
For the special cases of Euclidean projections \citep{deutsch1995dykstra} 
and Sinkhorn Balancing \citep{knight2008sinkhorn},
linear convergence has been proven. However we are unaware
of a linear convergence proof for general Bregman divergences. 

\item 
We hope that many of the techniques from the expert setting literature 
can be adapted to combinatorial online learning.
This includes 
lower bounding weights for shifting comparators \citep{herbster1998tracking}
and sleeping experts \citep{bousquet2002tracking,adamskiy2012putting}. 

\item In this paper, we studied the online learning problem in \emph{full information} setting,
where the learner receives the loss of its choice in such a way that the loss of any of 
the possible solution can be easily computed.
In the \emph{bandit} setting, however, the learner only observes the loss it incurs.
In the multipath learning problem, this means that the learner only observes
the loss of its predicted multipath and the losses on the multiedges are not revealed.
The algorithms in bandit settings usually apply EH or CH over the \emph{surrogate} loss vector
which is an unbiased estimation of the true unrevealed loss vector 
\citep{cesa2012combinatorial, gyorgy2007line, audibert2013regret, audibert2011minimax}.
Extending our methods to the bandit settings by efficiently computing the surrogate loss vector
is a potentially fruitful future direction of this research.

\item 
\textit{Online Markov Decision Processes (MDPs)} 
\citep{even2009online, dick2014online} is an online
learning model that focuses on the sequential revelation 
of a solution using a sequential state based model.
This is very much related to learning paths and the
sequential decisions made in our dynamic programming framework.
Connecting our work with the large body of research on MDPs
is a promising direction of future research.
\item
There are several important dynamic programming
instances that are not included 
in the class considered in this paper:
The Viterbi algorithm for finding the most probable path in a graph,
and variants of Cocke-Younger-Kasami (CYK) algorithm for
parsing probabilistic context-free grammars.
The solutions for these problems are min-sum type
optimization problem after taking a log of the
probabilities. However taking logs creates unbounded losses.
Extending our methods to these dynamic programming problems
would be very worthwhile.

\end{itemize}

\bibliographystyle{plain}
\bibliography{online_dp}

\end{document}